\pdfoutput=1
\RequirePackage{fix-cm}
\documentclass[twocolumn]{svjour3}          
\smartqed  
\usepackage{graphicx}
\usepackage{comment}
\usepackage{amsmath,amssymb} 
\usepackage{xcolor}
\usepackage{epsfig}
\usepackage{subfigure}
\usepackage{multirow}
\usepackage{setspace}
\usepackage[ruled,vlined]{algorithm2e}

\newcommand{\bp}[1]{{\color{black}#1}}

\newcommand{\ye}[1]{{\color{black}#1}}
\newcommand{\yep}[1]{{\color{black}#1}}
\begin{document}

\title{Efficient Joint-Dimensional Search with Solution Space Regularization for Real-Time Semantic Segmentation
}


\author{Peng Ye $^\dagger$  \and
        Baopu Li $^\dagger$ \and
        Tao Chen \textsuperscript{*} \and
        Jiayuan Fan  \and
        Zhen Mei  \and
        Chen Lin  \and
        Chongyan Zuo  \and
        Qinghua Chi  \and
        Wanli Ouyang
}


\institute{Peng Ye ($^\dagger$--co-first author) \and Tao Chen (*--corresponding author) \and Jiayuan Fan \and Zhen Mei \at
          School of Information Science and Technology, Fudan University, Shanghai, China \\
        \email{eetchen@fudan.edu.cn}
        \and
        Baopu Li ($^\dagger$--co-first author) \at Baidu Research  \\
        \and Chongyan Zuo \and Qinghua Chi 
         \at Huawei Inc. China\\
         \and Chen Lin 
         \at University of Oxford, England\\
         \and Wanli Ouyang 
         \at University of Sydney, Australia; Shanghai AI Laboratory \\
}

\date{Received: date / Accepted: date}

\maketitle
\begin{abstract}
Semantic segmentation is a popular research topic in computer vision, and many efforts have been made on it with impressive results. In this paper, we intend to search an optimal network structure that can run in real-time for this problem. Towards this goal, we jointly search the depth, channel, dilation rate and feature spatial resolution, which results in a search space consisting of about \ye{$2.78\times10^{324}$} possible choices. To handle \ye{such a} large search space, we \ye{leverage} differential architecture search methods. However, the architecture parameters searched using existing differential methods need to be discretized, which causes the discretization gap between the architecture parameters found by the differential methods and their discretized version as the final solution for the architecture search. Hence, we relieve the problem of discretization gap from the innovative perspective of solution space regularization. Specifically, a novel Solution Space Regularization (SSR) loss is first proposed to effectively encourage the supernet to converge to its discrete one. Then, a new Hierarchical and Progressive Solution Space Shrinking method is presented to further achieve high efficiency of searching. In addition, we theoretically show that the optimization of SSR loss is equivalent to the $L_{0}$-norm regularization, which accounts for the improved search-evaluation gap. Comprehensive experiments show that the proposed search scheme can efficiently find an optimal network structure that yields an extremely fast speed (175 FPS) of segmentation with \ye{a small model size} (1 M) while maintaining comparable accuracy. 

\end{abstract}

\section{Introduction}
\label{intro}
As a fundamental problem in computer vision, semantic segmentation predicts pixel-level labels for images. Thanks to the strong ability of convolutional neural networks (CNNs)~\cite{qu2020residual,hu2019graph,liang2018real}, many works~\cite{chen2017rethinking,zhao2017pyramid} have achieved encouraging performance on benchmarks~\cite{cordts2016cityscapes,zhou2019semantic,yu2018bdd100k}. However, these high-quality segmentation models usually require tremendous computation and memory overhead, which conflicts with the real-world resource-constrained devices. To tackle this issue, recent works~\cite{romera2017erfnet,li2019dfanet,yu2018bisenet} focus on designing low-latency, light-weight and accuracy-acceptable segmentation networks. 
For achieving this goal, more expertise and human efforts are inevitably needed to design optimal network structure and reduce the computation under limited resources as much as possible.


    \begin{figure*}[t]
      \centering
      \includegraphics[width=6.2in]{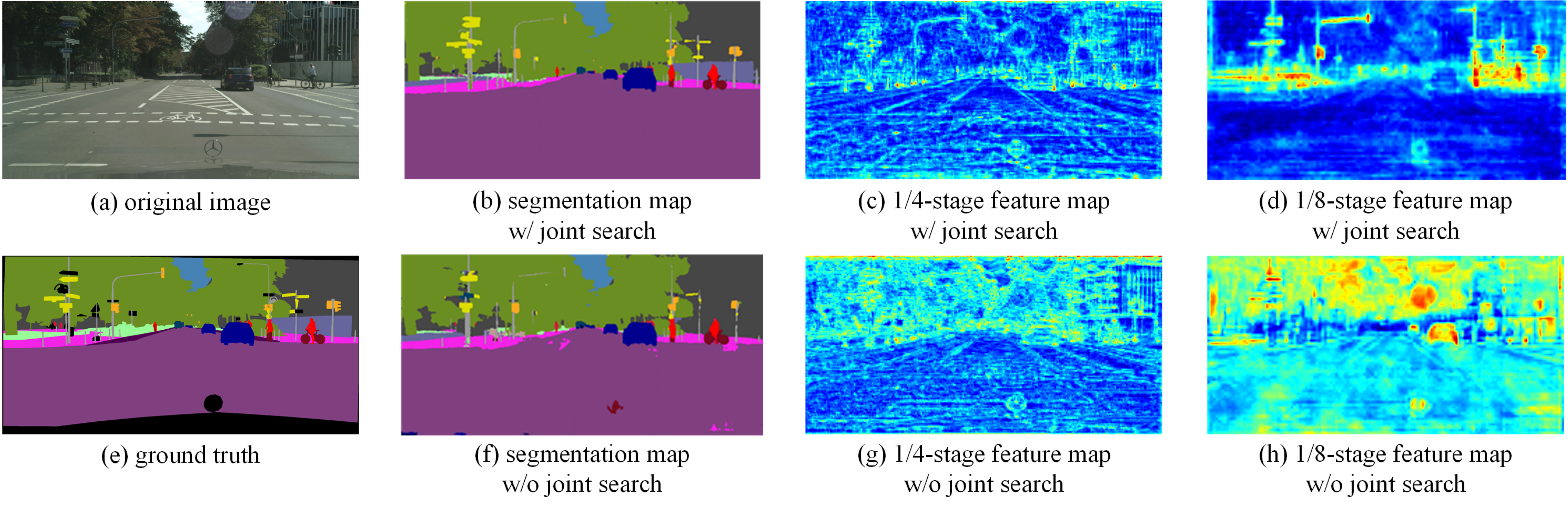}
      \caption{The importance of contextual information and spatial details. (a) and (e) denotes the input image and ground truth. (b)-(d) denotes the output of traditional model. (f)-(h) denotes the output of the model obtained by joint-dimensional search.
    }
      \label{fig:search_space}
      \vspace{-12pt}
    \end{figure*}


As an emerging method, neural architecture search (NAS) that optimizes both accuracy and resource utilization turns out to be a good candidate for this challenging trade-off. In particular, one-shot NAS that applies the weight sharing principle for the supernet and subnet has received a lot of attention due to its high efficiency. Along this direction, lots of researchers have made great efforts for NAS-based segmentation methods~\cite{liu2019auto,zhang2019customizable,lin2020graph,chen2019fasterseg,sun2021real}. As a pioneer, Auto-DeepLab~\cite{liu2019auto} searches the shared cells and the network-level path for high-quality segmentation. Recently, some works~\cite{zhang2019customizable,lin2020graph,chen2019fasterseg,sun2021real} utilize resource constraints to find real-time segmentation network. Specially, CAS~\cite{zhang2019customizable} searches the multi-scale decoder cell, GAS~\cite{lin2020graph} explores the cell-level diversity, and FasterSeg~\cite{chen2019fasterseg} integrates the multi-resolution branches into search, while AutoRTNet~\cite{sun2021real} jointly searches the network depth, the downsampling strategy, and an aggregation cell. Compared to human-designed networks, these methods appear to be more competitive and illustrate promising accuracy. However, these works simply implement segmentation task with NAS, using search space similar to that of classification tasks~\cite{liu2018darts}, or searching a pattern that has been identified useful on handcrafted networks. One pivotal point for semantic segmentation, that is, how to preserve more contextual information and spatial details, is seldomly considered. As shown in Fig.~\ref{fig:search_space}, the traditional segmentation model cannot handle well on the detailed regions such as the traffic signs in the road and the wall behind the person (in the middle of the right side of the segmentation map). After applying the model searched on different dimensions, more contextual information (e.g., person) and spatial details (e.g., edge) are extracted. The corresponding feature map visualization shows more distinct highlight in discriminative objects and regions and has clearer contours and edges\ye{. Thus,} the difficult regions described above can be segmented successfully.

For extracting more contextual and spatial features for semantic segmentation mentioned above, some important factors such as the network depth, dilation rate, feature spatial resolution and channel number should be comprehensively considered at the same time. Specifically, we first propose a multi-dimensional search space that jointly considers the above four factors simultaneously. However, such a multi-dimensional search may cause \textit{a huge search space} with different fine-grained levels. For example, when the above four dimensions are jointly searched, the search space is about $2.78*e^{324}$ (Please refer to the experimental section for the derivation process of this large number). Thus, we turn to differential architecture search methodology due to its high search efficiency~\cite{liu2018darts}. Then, we further apply continuous relaxation to conduct the differential joint-dimensional search. But the problem of \textit{discretization gap} which may greatly affect the performance of the searched subnetwork remains to be solved, and such a problem can be evidenced by Fig.~\ref{fig:different_layer_prob} and Fig.~\ref{fig:gap} in our experiments. Discretization gap is caused by the architecture parameter discretization process while finishing search in continuous architecture space, which will further lead to that the performance of the supernet cannot well indicate the performance of the final searched subnet. On multi-dimensional search of segmentation task, this issue will be further exacerbated because the search space is larger and more complex. 

To overcome the above two problems, namely huge search space and discretization gap in differential search, we propose to \ye{gradually optimize the continuous architectural distribution towards the discrete solution} from the novel perspective of solution space regularization (SSR). In particular, by considering each search choice for different architectures as a possible probability, we first design a SSR loss that can minimize the product of all the probabilities, to push candidates with approximate weights far from each other and make the learned supernet closer to the final discrete one. Such an effort can effectively mitigate the discretization gap problem that is common to differential related methods. In addition, it may also speed up the searching process and make the search process easier to find the optimal solution. More importantly, we give a theoretical analysis that the optimization of SSR loss is equivalent to the $L_{0}$-norm regularization, which accounts for the minimized discretization gap. Secondly, to strengthen SSR loss and \ye{reduce} the adverse impact of insignificant candidates, a hierarchical and progressive solution space shrinking strategy is proposed to further reduce the discretization gap and speed up the search process. Both the SSR loss and the hierarchical and progress shrinking can effectively overcome the aforementioned two issues, leading to an efficient searching process for the optimal segmentation model.
\ye{The contributions of our work can be summarized as the following:}
\begin{itemize}
  \item We design an efficient multi-dimensional search framework to jointly search the network architecture of depth, dilation rate, feature spatial resolution and channel for semantic segmentation, from the innovative perspective of Solution Space Regularization.
  \item A novel Solution Space Regularization (SSR) loss is advanced to encourage the architecture parameters converging to extreme values of selection/deselection, making the learned continuous architecture parameters closer to the final discrete ones for deciding network architecture. \bp{Theoretically}, we show that the optimization of SSR loss is equivalent to \bp{$L_{0}$-norm} regularization, from different standpoints. Experimentally, we show that the intractable discretization gap problem can be well handled, even \ye{on the} joint-dimensional search of segmentation.
  \item A new hierarchical and progressive solution space shrinking strategy is proposed to effectively reduce the solution space and improve the search efficiency on multi-dimensional search space.
\end{itemize}
We achieve extremely fast speed, light-weight parameters and FLOPs, while maintaining competitive accuracy on three benchmark datasets, that is, Cityscapes, Camvid and BDD. For example, the searched model can yield a performance of 11.0G FLOPs, 1.0M parameters, 175 FPS and 72.6\% mIoU on Cityscapes test set.

\section{Related Works}
\subsection{Semantic Segmentation}
\subsubsection{Manual Designed Model.}
High-quality segmentation algorithms can achieve pretty promising performance nowadays. In terms of real-time segmentation algorithms, there are two mainstreams. One-branch approaches construct relatively lighter backbones with some lightweight operations~\cite{paszke2016enet,romera2017erfnet} or introduce efficient feature aggregation modules ~\cite{li2019dfanet,orsic2019defense}. ERFNet ~\cite{romera2017erfnet} utilizes factorized residual layers to build an efficient network. DFANet ~\cite{li2019dfanet} designs sub-network and sub-stage aggregation modules via a lightweight backbone. Multi-branch approaches speed up the inference with multi-path. ICNet ~\cite{zhao2018icnet} adopts multi-resolution image cascade. BiSeNet~\cite{yu2018bisenet} and BiSeNetV2~\cite{yu2020bisenet} fuses a spatial path and a context path. These handcrafted algorithms provide important insights on how to design an efficient architecture for image semantic segmentation. In contrast, we aim to make use of search based algorithm to automatically find the optimal network for semantic segmentation.
\subsubsection{NAS-based Model.}
There have been some works using NAS methods on segmentation tasks. Auto-DeepLab ~\cite{liu2019auto} is the pioneering work in this direction that searches the cell-level structure and the network-level downsampling rates. CAS~\cite{zhang2019customizable} searches different types of cells under resource constraints. GAS~\cite{lin2020graph} releases the cell-sharing restrictions and uses a graph convolution network to guide the search process. FasterSeg~\cite{chen2019fasterseg} integrates multi-resolution branches into the search space and conducts teacher-student co-searching. AutoRTNet~\cite{sun2021real} jointly decides an aggregation cell, the network depth and downsampling strategy. These methods can explore more possibilities and achieve better trade-off between accuracy and resource constraints. However, the key problem of segmentation task, how to \ye{extract} more contextual information and spatial details with limited computational resources, has not been explored. In this paper, we attempt to utilize joint-dimensional search to effectively address this problem. 

\subsection{Neural Architecture Search}
Neural architecture search (NAS) has the ability to automate the architecture design process, thus \bp{receiving a lot of attention. Many}  novel methods have been proposed with impressive results, we only briefly review some typical and directly related methods here.
\subsubsection{Multi-dimensional Search.}
OFA~\cite{cai2019once} and BigNAS~\cite{yu2020bignas} can train a multi-dimensional supernet of elastic depth, channels, kernel size, and input resolution. However, they focus on using distillation technologies to train a supernet well and may not be efficient in searching. Though FBNetV2~\cite{wan2020fbnetv2} can search the network architecture of depth, channels, kernel size, and input resolution, it ignores the important search space of dilation and feature spatial resolution, which may play important roles for semantic segmentation. \ye{Moreover, our search method is rooted in solution space regularization, which is orthogonal to and can be utilized to improve the search of FBNetV2. EfficientNet~\cite{tan2019efficientnet} proposes a model scaling method that balances network depth, width and input resolution for image classification, while our method jointly searches the dilation rate, spatial resolution, depth and channel for semantic segmentation. Besides, the above methods cannot be directly transferred from classification task to segmentation task, because the latter needs more fine-grained selection of extracted features. SpineNet~\cite{du2020spinenet} searches a backbone with scale-permuted features and cross-scale connections for object detection. Different from SpineNet that uses scale-permuted features for spatial details preservation, our method employs spatial resolution search for capturing more contextual information and reducing the computation cost.} CRNAS~\cite{liang2019computation} realizes two-dimensional computation allocation for detection, but the stage and dilation allocation are separately conducted by one-shot method and greedy algorithm. In this work, we propose the differential four-dimensional co-searching for semantic segmentation \ye{via} the new perspective of solution space regularization.

\subsubsection{Deterministic Search.}
SNAS~\cite{xie2018snas} \ye{stochastically} samples architecture encoding and utilizes the Gumbel softmax method to update the search gradient. However, due to the sampling property, SNAS may not efficiently handle large and complex search space that is the typical characteristic to model the joint distribution of different search dimensions and suffer from the discretization gap problem. Whereas, the proposed method can be used to improve such effects of SNAS. FairDARTS~\cite{chu2019fair} introduces an auxiliary MSE loss to push architectural weights towards 0 or 1. However, the \ye{resulting} multiple solutions may not be suitable for joint-dimensional search, which expects one optimal choice for each dimension. To address these problems, we propose novel \bp{solution space regularization (SSR)} based approaches to push continuous architectural distribution gradually towards the discrete one. The method of solution space regularization is complementary to the above works.
\subsubsection{Progressive Search.}
PDARTS~\cite{chen2021progressive} drops a fixed number of candidates of each edge in each stage while manually \ye{enlarging} the search depth for cells. SGAS~\cite{li2020sgas} picks out the edge that has the best discriminability and certainty in all stages based on a few criteria. However, both the above progressive search methods are only used for operator level but ignore that \bp{the inter-level and intra-level choice may be both important in a search space with different fine-grained levels}. Therefore, we propose a hierarchical and progressive shrinking strategy for joint-dimensional search. In addition, they still suffer from the discretization gap problem, which is the major focus of this work.

	\begin{figure*}[t]
		\centering
		\includegraphics[width=0.95\linewidth]{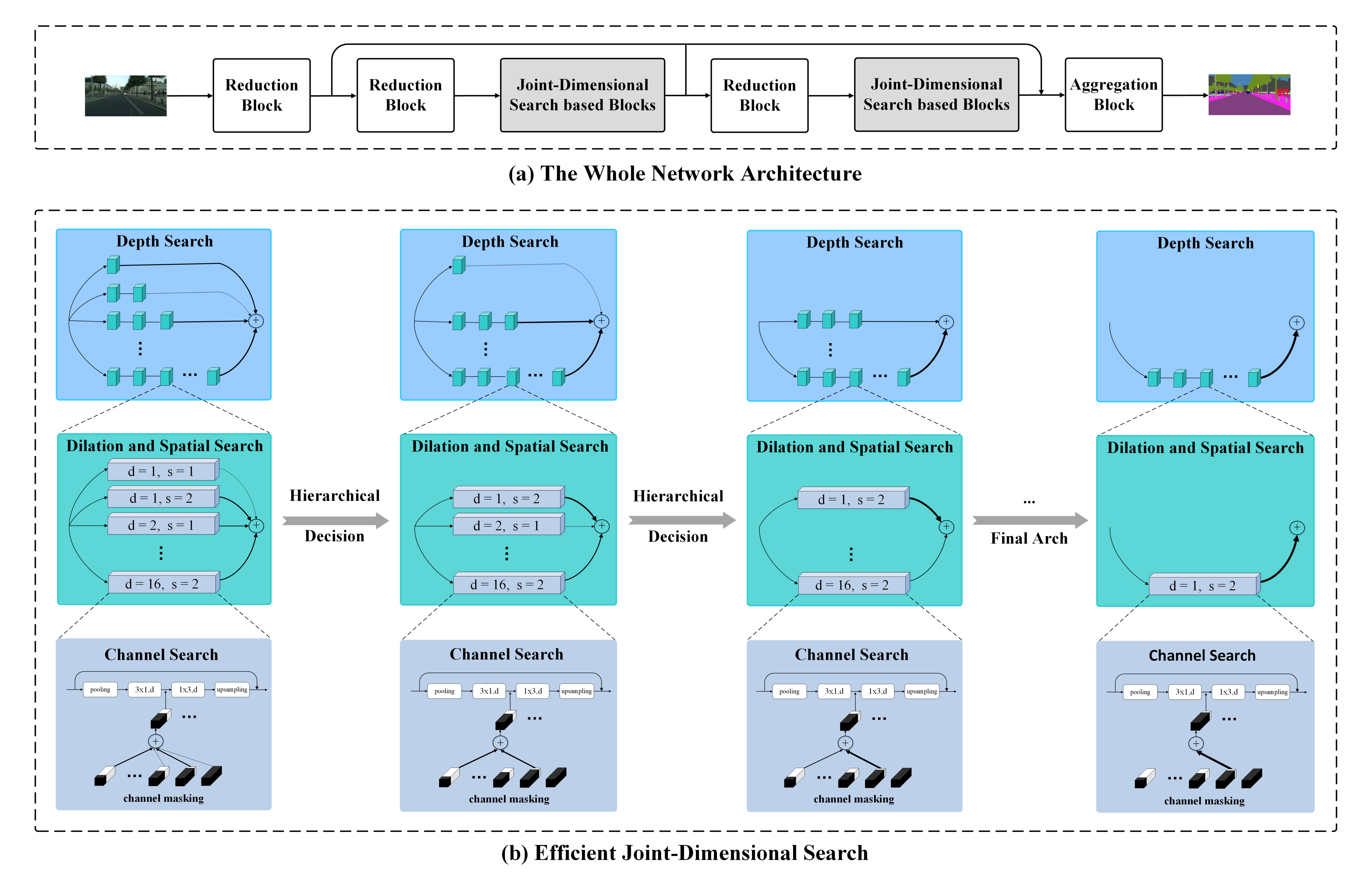}
		\caption{The overall framework of the proposed efficient joint-dimensional search framework for semantic segmentation. (a) The whole network architecture. We search the common three-stage segmentation backbone separated by reduction blocks used in~\cite{romera2017erfnet}, and decode the extracted features with an aggregation block. (b) Joint-dimensional search for depth, dilation rate, spatial resolution, and channel. }
		\label{fig:pipeline}
		\vspace{-12pt}	
	\end{figure*}

\section{Proposed Method}
In this section, we first detail the multi-dimensional search space for semantic segmentation, followed by a continuous relaxation for differentiable search. We then present the efficient joint-dimensional search framework, Solution Space Regularization (SSR) Search. Finally, we describe how to speed up the joint-dimensional search by the novel SSR loss and hierarchical and progressive solution space shrinking strategy. The pipeline of our efficient joint-dimensional search for semantic segmentation is shown in Fig.~\ref{fig:pipeline}.
        
\subsection{Multi-Dimensional Search Space}
To jointly search the network architecture of depth, channel, dilation rate, and feature spatial resolution so as to achieve better contextual information and spatial details preservation for semantic segmentation, we can formalize the joint-dimensional search problem as:
    \begin{equation} \label{eqn1}
    \min\limits_{A\in Arch(D_{e},D_{i},S,C)}L_{val}(w, A) 
    \end{equation}
where $Arch(D_{e},D_{i},S, C)$ denotes the four-dimensional (i.e. Depth,  Dilation rate, Spatial resolution, Channels) search space, $L_{val}$ is the validation loss. Considering the consumption of computational resources and time, we intend to solve this objective in a differentiable manner. We first present their respective search space in the following paragraphs, and the specific configuration of each dimension is shown in Table~\ref{tab:macro} and Table~\ref{tab:micro}.

\bigskip\noindent\textbf{Depth Space.} The backbone network usually has several stages, which output intermediate features with increasing downsampling rates. FPN~\cite{lin2017feature} has shown that the shallower stage generates features with more spatial detail information, while the deeper stage yields features with more semantic information. CRNAS~\cite{liang2019computation} has indicated that the number of layers (i.e., depth) of each stage has a great impact on its effective receptive field. Inspired by these works, we enable the depth search within the segmentation backbone. Fig.~\ref{fig:pipeline}(a) shows our whole network architecture, we search the depth on two stages of the common segmentation backbone. The upper part of Fig.~\ref{fig:pipeline}(b) shows our depth search space, which contains several branches with different numbers of layers. To \ye{relieve} computational burden, we introduce the weight sharing strategy, through which all the other branches shown in the block with blue background color inherit weights from the branch with the maximum number of layers. 

\bigskip\noindent\textbf{Dilation Rate and Spatial Resolution Space.} Dilated convolution has the ability to enlarge the receptive field without extra parameters or computations, thus it is popular in semantic segmentation. ~\cite{chen2017rethinking} and ~\cite{wang2018understanding} have shown that different permutations and combinations of dilated convolutions will greatly impact the performance and the gridding problem. Previous manually designed methods finitely explore the possibilities, thus solving this problem with NAS-based methods is important. Spatial resolution search ~\cite{liu2020shape,guo2020multi} has been identified \ye{as} useful to reduce the spatial redundancy of feature maps in image recognition. Referring to ~\cite{guo2020multi}, we sequentially process the input feature map with $s \times s$ average pooling, some operators and bilinear upsampling. This special design brings another benefit that $s$ times pooling implies $s$ times larger receptive field, which sometimes is indispensable for segmentation. Since both the dilation rate and spatial resolution directly affect the input feature map of each layer, we integrate them together in the same level search space. As shown in the middle part of Fig.~\ref{fig:pipeline}(b), our search space of each layer is made up of the \textit{combinations} of different dilation rates and feature spatial resolutions. For simplicity, we denote it as dilation-and-spatial search space.

\bigskip\noindent\textbf{Channel Space.} Most methods of channel search are specialized for image recognition, and they may not work well for more complex tasks like object detection and semantic segmentation. One possible reason could be that the latter is more sensitive to not only semantic information but also spatial details, while most channel search methods ignore the influence of other dimensions when searching. To mitigate this effect, we jointly search the channel dimension and other dimensions to achieve better trade-off between semantic and spatial information. Our channel search space is shown in the bottom part of Fig.~\ref{fig:pipeline}(b), and we define the basic operator as the simplified version of the efficient residual factorized convolution in ~\cite{romera2017erfnet}, which consists of a set of $3\times 1$ and $1\times 3$ convolutions. And we search the channel number of $3\times 1$ and $1\times 3$ convolutions independently. Moreover, we extend the channel masking strategy proposed in ~\cite{wan2020fbnetv2} to realize more convenient channel search. Specifically, the channel masking after each convolution layer is the weighted sum of several zero-one masks, and all the channel masking in the search space of the same layer share the same set of zero-one masks. 

\subsection{Continuous Relaxation for Differential Search}
To enable differential search in the four-dimensional search space $Arch(D_{e},D_{i},S, C)$, we conduct continuous relaxation in different dimensions step by step in the following. To facilitate the depth search as shown in Fig.~\ref{fig:pipeline}(b), we first obtain a weighted \bp{sum} of the layer outputs of different depths in each stage, and then utilize it as the input for the next stage, denoted as: 
\begin{equation} \label{eqn2}
T_i = \sum\limits_{\alpha_j^i} \sigma(\alpha_j^i)
l_j^i,\quad \alpha_j^i\in Arch(D_{e})
\end{equation}
where $l_j^i, \alpha_j^i $ and  $\sigma(.)$ are the \bp{output} of candidate depth $j$ in stage $i$, the corresponding depth architecture parameter, and the activation function, respectively. 

For dilation rate and feature spatial resolution search, as shown in  Eq.~(\ref{eqn3}), each layer is the weighted \bp{sum} of the output of different operators: 
\begin{equation} \label{eqn3}
  \begin{split}
  &L_{i,j} = \sum\limits_{\beta^{i,j}_{r,s}} 
  \sigma(\beta^{i,j}_{r,s}) o^{i,j}_{r,s}, \\
  &\beta^{i,j}_{r,s}\in Arch(D_{i},S)
  \end{split}
\end{equation}
where $o^{i,j}_{r,s},\beta^{i,j}_{r,s} $ denote the operator with dilation rate $r$ and spatial \bp{resolution} $s$, and the corresponding architecture parameter respectively.
\bp{Specifically}, we formulate the operator structure as:
\begin{equation} \label{eqn4}
o^{i,j}_{r,s} = up(\;op^{i,j}_{r}(pool^{i,j}_{s\times s}(l_{j-1}^i)))  
\end{equation}
where $pool^{i,j}_{s\times s}$, $op^{i,j}_{r}$, $up$ mean $s\times s$ average pooling, convolution operators with dilation rate $r$ and bilinear upsampling respectively.

In addition, we search the output channel number of each convolution, and this process is defined as:
\begin{equation} \label{eqn5}
  \begin{split}
    &op^{i,j}_{r} = conv \circ \sum_{\gamma_k^{i,j,r,s}} 
    \sigma(\gamma_k^{i,j,r,s}) M_k^{i,j,r,s},   \\
    &\gamma_k^{i,j,r,s}\in {Arch(C)}
  \end{split}
\end{equation}
where $\gamma_k^{i,j,r,s} $ means the architecture parameter for output channel search of each convolution. $M_k$ denotes the shared zero-one masks, conv is the typical convolutional operation, and $\circ$ is weighted multiplication. 

Note that, all $\alpha$, $\beta$ and $\gamma$ are activated with sigmoid activation ($\sigma$) and normalized. They jointly encode the \bp{multi-dimensional search space}  and form $A$ in Eq.~(\ref{eqn1}). In differentiable search, solving the problem of Eq.~(\ref{eqn1}) involves a bi-level optimization problem.
\begin{equation} \label{eqn6}
  \begin{split}
    &\mathop{\min}_{A} L_{A}(w^{\ast}(A), A), \\
    &s.t.\ w^{\ast} = \arg\mathop{\min}_{w}L_{w}(w,A)
  \end{split}
  \vspace{-6pt}
\end{equation}
where $L_w$ and $L_A$ denote the validation loss functions of network weights $w$ and architecture parameters $A$ respectively. Referring to~\cite{liu2018darts}, we solve this problem by alternatively updating $w$ and $A$ via gradient descent.

\subsection{Solution Space Regularization Search}
Though the above multi-dimensional search space is desirable for semantic segmentation, it may be challenging to optimize and most previous differentiable methods cannot handle such a large and complex search space. Specifically, most differentiable methods first optimize the architectural parameter distribution via gradient descent, then discretize the best architecture by selecting the candidate with the highest probability for every mixed operation and retrain it from scratch. \ye{However, most methods not only cannot guarantee that the parameter distribution is sufficiently distinguished, but also make decisions in a rather simple manner such that local candidates with approximate weights are not sufficiently considered, as will be shown in Fig.~\ref{fig:different_layer_prob} and Fig.~\ref{fig:gap}. Thus, these methods may suffer from serious discretization gap problem, that is, the architecture parameters learned by the differential methods and their discretized version as the final solution for the architecture search are different.} For \ye{the} joint-dimensional search of segmentation task, the discretization gap problem will be further exacerbated because the search space is larger and more complex while segmentation task is sensitive to the architecture selection of different \ye{dimensions}. In addition, the search process needs more consumption of time and \ye{resources} since segmentation task needs to process features of higher resolution than classification. To alleviate these problems, we present the innovative concept of solution space regularization. 

\begin{algorithm*}[t] \label{agm1}
\caption{Solution Space Regularization based Joint-Dimensional Search}
Create multi-dimensional architectural parameters $A=\{\alpha, \beta, \gamma\}$ and supernet weights $w$\;
Create depth-, dilation-and-spatial- and channel-level solution space with $\alpha$, $\beta$ and $\gamma$ respectively\;
\While{not the discrete solution}{
   1. update remaining architectural parameters A by descending $\nabla_{A} (L_A(w, A)+\lambda L_{SSR})$\;
  2. update unremoved network weights $w$ by descending $\nabla_{w} L_w(w, A)$\;
   ~~~(only the weights of unpruned candidates of each level needed to be kept and updated)\;
  3. \For{{\rm solution space in [depth-level, dilation-and-spatial-level, channel-level]}}{
  \If{{\rm Eq.~(\ref{eqn13}) is satisfied}}{
  shrink solution space by removing architectural parameters and supernet weights from $A$ and $w$\;}}}
\end{algorithm*}


\bigskip\noindent\textbf{Motivation.}
Owing to continuous relaxation, the task of architecture search is reduced to the optimization of continuous architectural parameters, thus it can be realized by solving Eq.~(\ref{eqn6}). However, each coin has its two sides and continuous relaxation is not an exception. As shown in Fig.~\ref{fig:ssr}(a), the discrete search space is converted to the continuous \textit{solution space} (that is, any learned continuous architectural parameters can be regarded as a solution) because of continuous relaxation. For example, three candidates in search space can form a triangular plane of solution space, which contains infinite solutions. As illustrated in Fig.~\ref{fig:ssr}(b), the probability of each candidate is between 0 and 1, thus multiple locally optimal solutions may exist when optimizing, and the discrete solution obtained from the result of architecture optimization (black star) may differ from the real desired discrete solution (orange star). As such, previous differentiable methods can be considered as one kind of \textit{supernet solution optimization,} and the learned continuous architectural parameters (i.e., learned solution) may differ from the desired discrete solution. \ye{Thus, they may result in a} serious discretization gap problem and more consumption of time and computation. In fact, we hope that the final architectural parameter learned by Eq.~(\ref{eqn6}) should be equal to the desired discrete solution, that is, the continuous parameter distribution can be gradually optimized to the discrete representation as close as possible. This implies a sparse representation and reconstruction process. Inspired by the observation in Fig.~\ref{fig:ssr} and the above motivation, we introduce Solution Space Regularization (SSR) into the process of supernet solution optimization, making the learned architectural parameter closer to the discrete solution, and gradually shrinking the solution space. As such, the discretization gap problem can be effectively relieved and the consumption of computation and time can be greatly reduced.




\bigskip\noindent\textbf{Proposed Scheme.} To realize the target of \textit{optimizing the solution space toward the discrete solution}, we propose a novel solution space regularization loss and a new hierarchical and progressive solution space shrinking strategy (detailed in Subsection \ref{sec:ssr loss} and \ref{sec:sss}). The former method is used to ensure the optimized continuous architecture parameters come closer to the final discrete solution for discretization gap minimizing\ye{. The} latter approach is applied to gradually remove negligible solutions for time and resource saving. They work together to achieve architecture search effectively and efficiently. The procedure of the proposed joint-dimensional search framework is shown in Algorithm~\ref{agm1}. 

\subsection{Solution Space Regularization Loss} \label{sec:ssr loss}

Given a set of candidate architecture parameters, that is, $\alpha$, $\beta$ and $\gamma$ in Eq.~(\ref{eqn2}), Eq.~(\ref{eqn3}) and Eq.~(\ref{eqn5}), we denote the activated and normalized output \ye{of each dimensional search}  as $[p_1,...,p_i,...,p_n]$, where $p_1+p_2+...+p_n=1$. Ideally, we want to encourage the architecture parameters to converge to extreme values of selection/deselection. In other words, we wish the parameter distribution can be gradually optimized to the discrete one $[0,...,1,...,0]$ under the simplex constraint~\cite{tian2011note}. This problem can be formulated as: 
\begin{equation} \label{eqn9}
  \begin{split}
    & [p_1,...,p_i,...,p_n] \Rightarrow [0,...,1,...,0] \\
    & s.t.\sum_{i=1}^{n} p_i=1
  \end{split}
  \vspace{-6pt}
\end{equation}
where $\Rightarrow$ denotes the gradual optimization process.

    \begin{figure}[t]
      \centering
      \includegraphics[width=3.2in]{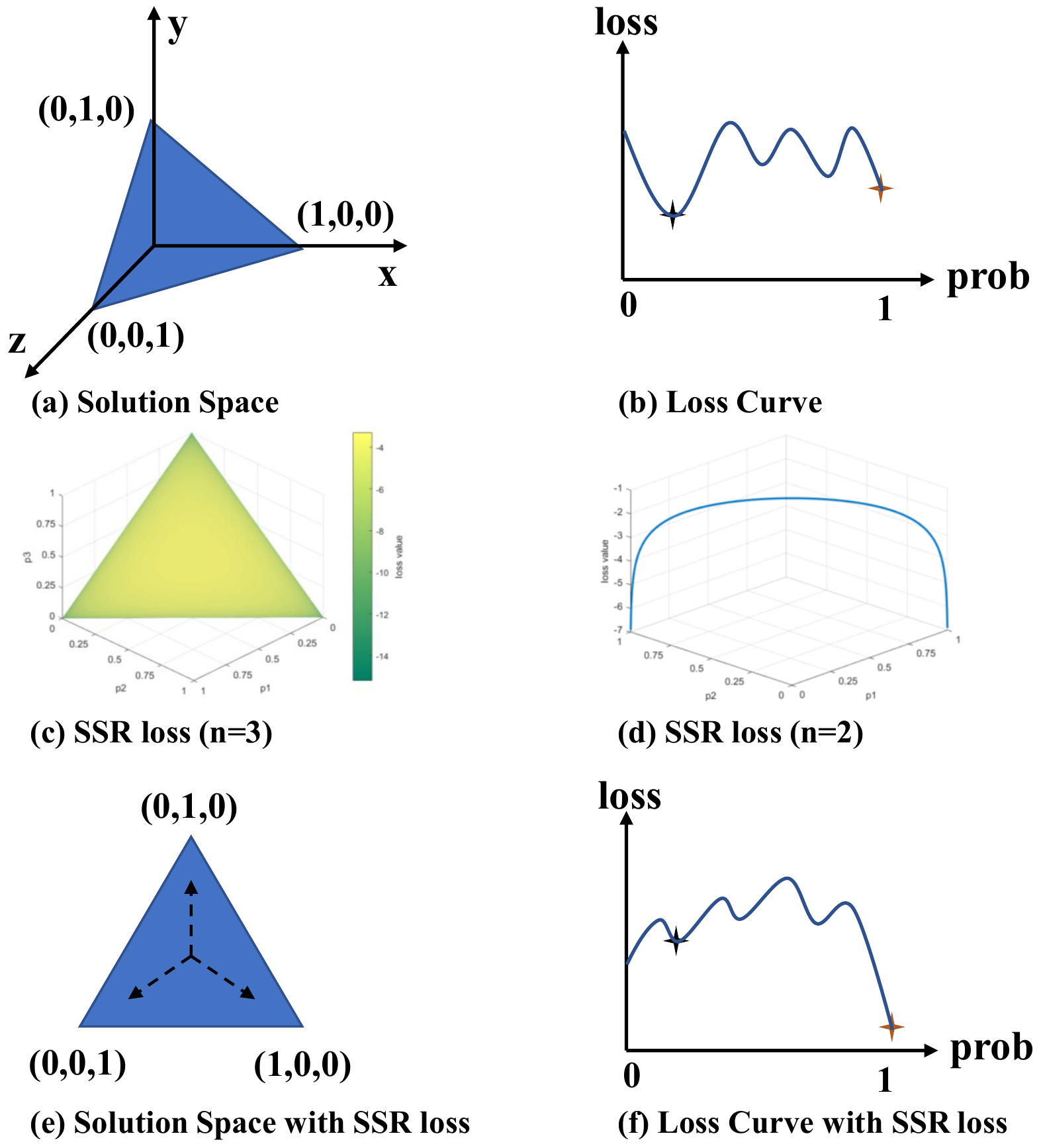}
      \caption{\yep{The importance of Solution Space Regularization (SSR) loss. (a) Original solution space. (b) Original searching loss curve relative to the probability of a certain candidate. (c)-(d) SSR loss when $n=3$ and $n=2$. (e) Solution space with SSR loss,  the arrow means that the learned solution inclines to different vertexes corresponding to the selection/dis-selection of different candidates. (f) Loss curve with SSR loss.}}
      \label{fig:ssr}
      \vspace{-6pt}
    \end{figure}

This is a \ye{mathematically multi-objective} constrained optimization problem with $n$ probability to be optimized, which has more than one single solution, and the interaction between multiple objectives makes the optimization process full of uncertainty. Based on the fact that we hope the parameter distribution can gradually change from $[0.333,0.333,0.333]$ to $[0.2,0.5,0.1]$ to $[0,1,0]$ when there are three candidates, which is a process of uncertainty ($U$) minimization, thus an alternative approximate solution can be formulated as: 
\begin{equation} \label{eqn10}
  \begin{split}
    & min \ U \Leftrightarrow min \ p_1 \cdot p_2 \cdot ... \cdot p_n \\
    & s.t.\sum_{i=1}^{n} p_i=1
  \end{split}
  \vspace{-6pt}
\end{equation}


Based on the above new formulation, we construct a novel loss function by applying a log function to each probability in Eq.~(\ref{eqn10}), so that we can transform the objective of multiplication into summation and facilitate the optimization process. Because the proposed loss can impose constraints to the solution space of architecture parameters, it is called Solution Space Regularization (SSR) loss, denoted as $L_{ssr}$:

\begin{equation} \label{eqn11}
L_{ssr} = \sum_{i=1}^{n}log(p_i)
\vspace{-8pt}
\end{equation}

When the number of candidates $n=3$ and $n=2$, the corresponding SSR loss is shown in Fig.~\ref{fig:ssr}(c) and Fig.~\ref{fig:ssr}(d), we can see that SSR loss becomes smaller when the architecture parameters come closer to the discrete solution $[0,...,1,...,0]$. Thus, when SSR loss is introduced to the process of supernet solution optimization, as shown in Fig.~\ref{fig:ssr}(e), it can reduce the optimal solution space from the whole triangular plane to its vertices, which contains finite solutions (i.e., the number of candidates $n$). As illustrated in Fig.~\ref{fig:ssr}(f), under the effect of SSR loss regularization, we can make sure the searching loss reaches its minimum value when the probability of each candidate approaches towards extreme values of 0/1, which corresponds to the optimal solution to be searched. As such, the SSR loss can relieve the problem of discretization gap and existing multiple local optimal solutions.

Note that, the number of architecture parameters $n$ will be gradually reduced via the shrinking strategy proposed below, which can further strengthen the gap reduction ability and the numerical stability of SSR loss. Specifically, after the candidates with low weights are removed via the shrinking strategy, the remaining candidates with approximated weights will be further compared via the SSR loss. Since Eq.~(\ref{eqn11}) will become smaller when the probability for each candidate is further apart, SSR loss has exactly the ability to accelerate the competition process. Another potential benefit behind such combined effects lies in that it may be more important to compare candidates with high weights rather than low ones during the search process. Consequently, we can avoid some interference caused by useless candidates and smoothly come close to the optimal architecture. 

As a result, by applying SSR loss to different levels (detailed in Subsection~\ref{sec:sss}) of the multi-dimensional search space, the total loss of architecture parameters in Eq.~(\ref{eqn6}) can be now reformulated as:
\begin{equation} \label{eqn12}
L_{A}+\lambda L_{SSR} = L_{A}+\sum_{l=1}^{L} \rho_l L_{ssr}^l
\vspace{-8pt}
\end{equation}
where $\rho_1, ..., \rho_L$ are used to control the weights of different levels of the multi-dimensional search space.

\subsection{Solution Space Shrinking} \label{sec:sss}

Generally speaking, keeping all candidates when optimizing the weights of supernet not only causes more consumption of search time and computation \ye{resources}, but also interferes with the optimization process. To remedy this, we propose a novel hierarchical and progressive solution space shrinking strategy, which consists of hierarchical shrinking and progressive shrinking.

\bigskip\noindent\textbf{\bp{Hierarchical} Shrinking.} Considering that different levels of solution space have different \ye{influences} on the search process, we introduce a hierarchical shrinking strategy. Specifically, we first decouple the whole solution space into different fine-grained levels via architecture parameters $\alpha$, $\beta$ and $\gamma$, then sequentially deal with the depth-level (via $\alpha$), dilation-and-spatial-level (via $\beta$) and channel-level (via $\gamma$) solution space based on mutually independent SSR loss and progressive shrinking strategy (proposed below). Once some candidates in the former level are removed, all the corresponding candidates in the subsequent levels are not considered anymore. Because the architecture parameters and network weights update of removed candidates will be stopped, the solution space will be gradually optimized, and the search efficiency will be further improved.

\begin{figure}[t]
      \centering
      \includegraphics[width=3.3in]{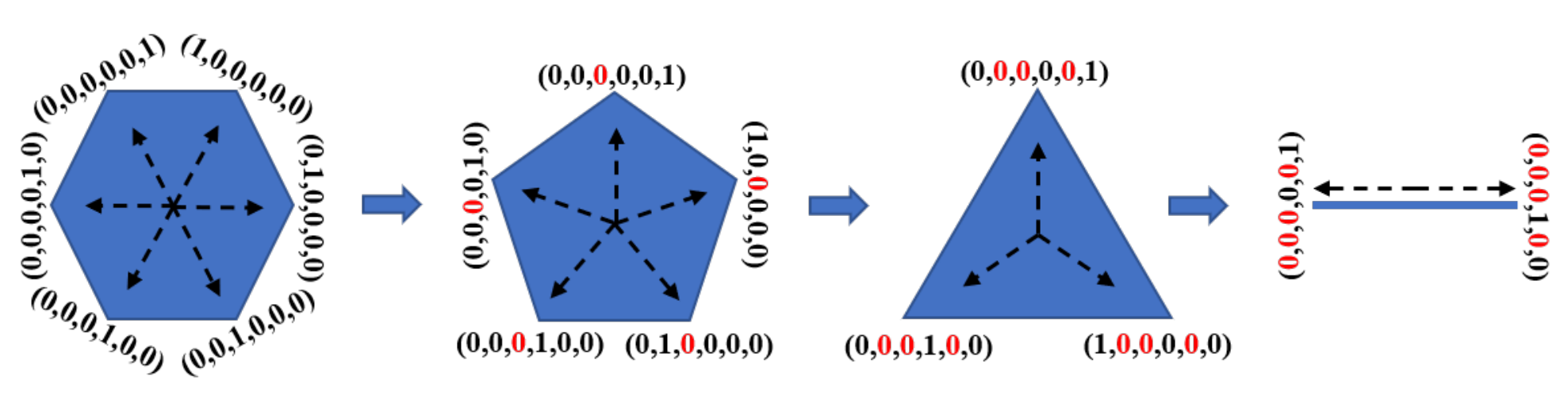}
      \caption{Illustration of the process of Solution Space Shrinking. The blue planes denote the solution space, and the arrow means that the learned solution inclines to different vertexes corresponding to the selection/dis-selection of different candidates. The probability vectors in different vertexes denote possible discrete solutions, and red 0 means the dis-selected candidates. \ye{With shrinking, the solution space will be gradually reduced to the final discrete solution.}
      }
      \label{fig:sss}
      \vspace{-6pt}
    \end{figure}

\bigskip\noindent\textbf{Progressive Shrinking.}
PDARTS~\cite{chen2021progressive} gradually remove the relatively weak operators for each edge, SGAS \cite{li2020sgas} progressively discretizes the most certain edge in all edges. However, different operators of different edges/ layers usually present different discriminability and certainty during the search, as will shown in Fig.~\ref{fig:different_layer_prob}. Based on this observation, we propose a level-based progressive shrinking method, which considers all operators of all edges/layers simultaneously and removes candidates with low enough architectural weights progressively. Considering the commonly used operator-level search space, we first define the retaining probability of each operator of each layer as:
\begin{equation} \label{eqn13}
  \begin{split}
  p^{i,j}_{m,n} = \frac{\sigma(\beta^{i,j}_{m,n})}{max{\left\{ \sigma(\beta^{i,j})  \right\}}} 
  \end{split}
\end{equation}
After that, for all candidates in the operator-level search space, a global threshold $h$ is used to remove some insignificant candidates. Once the condition $p^{i,j}_{m,n}<=h$ is satisfied, we directly discard the corresponding candidates and stop their architecture parameters and network weights update. With the assistance of SSR loss, the operator-level solution space will be gradually optimized to the discrete solution via progressive shrinking, as shown in Fig.~\ref{fig:sss}. Note that, we apply the proposed level-based progressive shrinking strategy to our different levels, namely, depth-level, dilation-and-spatial-level and channel-level, of the solution space and uniformly set $h=0.1$ to trade off the architecture discrepancy minimization and the search efficiency.

\section{Equivalence between SSR Loss and $L_{0}$ Norm }
We further present a theoretical analysis \ye{of} why the proposed search framework can realize the target of optimizing the solution space towards the discrete solution. Specifically, we demonstrate that the optimization of SSR loss is equivalent to the $L_{0}$-norm regularization from three different angles, with the help of the following p-norm-like function.
\begin{equation} \label{eqn14}
  \begin{split}
  E^{m}(p) = \sum_{i=1}^{n} {\left | p_i \right |}^m
  \end{split}
\end{equation}
when $m=0$, $E^{0}(p)={\left \| p \right \|}_0=\# (i \mid p_i \neq 0)$ denotes the standard \bp{$L_{0}$-norm} regularization term and $\#$ counts the total number of non-zero elements. 

\bigskip\noindent\textbf{\bp{Objective Equivalence.}} Considering that $E^{m}(p)$ is continuous and differentiable with respect to $m$:
\begin{equation} \label{eqn21}
  \begin{split}
  \frac{d E^{m}(p)}{dm}=\sum_{i=1}^{n} {\left | p_i \right |}^m \ln{\left | p_i \right |}
  \end{split}
\end{equation}
When $m\rightarrow0$, we obtain the following equality:
\begin{equation} \label{eqn22}
  \begin{split}
  \lim_{m\rightarrow0}\frac{d E^{m}(p)}{dm}=L_{ssr}(p)
  \end{split}
\end{equation}
Then, we can connect $L_{ssr}$ with $E^{m}(p)$ via performing first order Taylor series expansion at $m=0$:
\begin{equation} \label{eqn23}
  \begin{split}
  E^{m}(p) &= E^{0}(p)+m L_{ssr}(p)+o(m) \\
  &\approx E^{0}(p)+m L_{ssr}(p)
  \end{split}
\end{equation}
where $E^{0}(p)$ is a constant term. Thus as $m$ gets small, $E^{m}(p)$ begins to behave like $L_{ssr}(p)$.

\bigskip\noindent\textbf{\bp{Target Equivalence}.} In addition, we further show that $L_{ssr}$ and $E^{m}(p)\mid_{m\rightarrow0}$ have optimization target equivalence. Given the monotonically related functional $exp(L_{ssr}(p))=\prod_{i=1}^{n}p_i$ in Eq.~(\ref{eqn10}), we can relate it with $E^{m}(p)$ via the arithmetic-geometric mean inequality.
\begin{equation} \label{eqn18}
  \begin{split}
  (\prod_{i=1}^{n}{\left | p_i \right |}^m)^{\frac{1}{n}} \leqslant \frac{1}{n} \sum_{i=1}^{n}{\left | p_i \right |}^m
  \end{split}
\end{equation}
This implies for all $m$ and $p_i>0$, the following inequality is satisfied:
\begin{equation} \label{eqn19}
  \begin{split}
  (-\frac{1}{n}E^{(m^-)}(p))^{\frac{1}{m^-}}\leqslant [exp(L_{ssr}(p))]^{\frac{1}{n}}\leqslant (\frac{1}{n}E^{(m^+)}(p))^{\frac{1}{m^+}}
  \end{split}
\end{equation}
where $m^-\leqslant0$ and $m^+\geq0$. When taking the limit as $m\rightarrow0$, we get the following equality, thus build the connection between $L_{ssr}$ and $E^{m}(p)$.
\begin{equation} \label{eqn20}
  \begin{split}
  exp(\frac{1}{n}L_{ssr}(p)) = \lim_{m\rightarrow0}(\frac{1}{n}E^{m}(p))^{\frac{1}{m}}
  \end{split}
\end{equation}
We can see that the optimization target of $L_{ssr}$ is equivalent to that of $E^{m}(p)\mid_{m\rightarrow0}$.

\bigskip\noindent\textbf{\bp{Gradient Equivalence.} }From the gradient point of view, we can also demonstrate that $L_{ssr}$ and $E^{m}(p)\mid_{m\rightarrow0}$ have gradient equivalence when optimizing. Suppose that $m \neq 0$, the gradient of $E^{m}(p)$ with respect to $p_i$ can be represented as:
\begin{equation} \label{eqn15}
  \begin{split}
   \nabla_{p_i} E^{m}(p) = {\left | m \right |}{\left | p_i \right |}^{m-2}p_i
  \end{split}
\end{equation}
then the gradient vector of $E^{m}(p)$ can be written as the following factored representation:
\begin{equation} \label{eqn16}
  \begin{split}
   \nabla_{p} E^{m}(p) = \omega(p)\Pi(p)p
  \end{split}
\end{equation}
where $\omega(p)={\left | m \right |}$ and $\Pi(p)=diag({\left | p_i \right |}^{m-2})$.
Similarly, since we know that $0<p_i<1$, the gradient vector of $L_{ssr}$ can be represented as:
\begin{equation} \label{eqn17}
  \begin{split}
   \nabla_{p} L_{ssr}(p) = \omega_{ssr}(p)\Pi_{ssr}(p)p
  \end{split}
\end{equation}
where $\omega_{ssr}(p)=1$ and $\Pi_{ssr}(p)=diag(\frac{1}{p_i^2})$.
When we ignore the influence of constant term with respect to the minimization process, we can see that $\Pi(p)=\Pi_{ssr}(p)$ as we take $m=0$.

In summary, we relate $L_{ssr}(p)$ with $E^{m}(p)\mid_{m\rightarrow0}$ from different theoretical perspectives. As $E^{m}(p)\mid_{m\rightarrow0}=E^{0}(p)={\left \| p \right \|}_0$, we can say that $L_{SSR}$ minimization is equivalent to $L_{0}$-norm minimization, and the optimization problem in Eq.~(\ref{eqn6}) can be written as follows:
\begin{equation} \label{eqn24}
  \begin{split}
    &\mathop{\min}_{A} (L_{A}+\lambda L_{SSR}) \Leftrightarrow \mathop{\min}_{A} (L_{A}+\lambda {\left \| A \right \|}_0)\\
    &s.t.\ w^{\ast} = \arg\mathop{\min}_{w}L_{w}(w,A)
  \end{split}
  \vspace{-6pt}
\end{equation}
As we all know, $L_{0}$-norm is used to measure the number of non-zero elements in a vector. Thus, minimizing $L_{0}$-norm is equivalent to minimizing the number of non-zero elements. Such sparse effect is exactly what we want in the search process, because the target of optimizing the architectural parameters toward the discrete solution needs the parameter vector only has one non-zero entry. Under the simplex constraint, the optimal solution of $L_{0}$-norm minimization is the discrete solution $[0,...,1,...,0]$. Thus, solving Eq.~(\ref{eqn24}) will naturally mitigate the discretization gap problem that is common to previous differentiable NAS methods since the learned solution come closer to the discrete one. Note that $L_{0}$-norm regularization is \bp{dis-continuous}, thus is challenging for minimization. The proposed $L_{ssr}$ is not only suitable for optimization but can achieve the same effect as {$L_{0}$-norm} regularization.

\section{Experiments}
\subsection{Datasets and Evaluation Metrics}
\textbf{Datasets.} To verify the effectiveness of the proposed joint dimensional search method, we evaluate on three popular benchmark datasets: Cityscapes, CamVid and BDD. The Cityscapes dataset contains 5000 high-quality pixel-level annotated images, in which 2975 images are for training, 500 images are for validation and 1525 images are for testing. These images have a resolution of $1024 \times 2048$ and belong to \ye{the pre-defined} 19 classes. For \ye{a} fair comparison, the testing set does not provide ground truth. The Camvid dataset contains 701 images, in which 367, 101, and 233 images are for the training, validation and testing respectively. These images have a resolution of $720 \times 960$ and total 11 semantic categories. The BDD dataset contains 8000 images, in which 7000 images are for training and 1000 images are for validation. These images have a resolution of $720 \times 1280$ and share the same 19 classes as utilized in Cityscapes.


\bigskip\noindent\textbf{Evaluation Metrics.} For a comprehensive comparison, we use mean Intersection over Union per class (mIoU), Frame Per Second (FPS), the number of model Parameters (Params) and Floating point of Operations (FLOPs) as the evaluation metrics.

\begin{table}[!t]
\begin{center}
\footnotesize
\caption{Macro architecture of the searched backbone. OPS denotes the operators to be searched. DB is the downsampler block in~\cite{romera2017erfnet}. 1-5 means the depth to be searched.}
\label{tab:macro}
\begin{tabular}{ccccc}
\hline\noalign{\smallskip}
Input Size & C$_{in}$ &  C$_{out}$ & Operator & Depth \\
\noalign{\smallskip}
\hline
\noalign{\smallskip}
512 $\times$ 1024  & 3 & 16 & DB & 1\\
256 $\times$ 512 & 16 & 64 & DB & 1\\
128 $\times$ 256 & 64 & 64 & OPS & 2\\
128 $\times$ 256 & 64 & 64 & OPS & 1-5\\
128 $\times$ 256 & 64 & 128 & DB & 1\\
64 $\times$ 128 & 128 & 128 & OPS & 5\\
64 $\times$ 128 & 128 & 128 & OPS & 1-5\\
\hline
\end{tabular}
\end{center}
\vspace{-12pt}
\end{table}

\begin{table}[!t]
\begin{center}
\footnotesize
\caption{Micro architectures of OPS. NB-1D denotes the simplified Non-Bottleneck-1D block in~\cite{romera2017erfnet}. C$_{3\times 1}$ and C$_{1\times 3}$ means the output channel number of $3\times 1$ and $1\times 3$ convolution, which will be searched independently.}
\label{tab:micro}
\begin{tabular}{ccccc}
\hline\noalign{\smallskip}
OPS & Dilation & Spatial & C$_{3\times 1}$ and C$_{1\times 3}$ \\
\noalign{\smallskip}
\hline
\noalign{\smallskip}
NB-1D  & 1 & [1,2] & (C$_{max}$-32):4:C$_{max}$ \\
NB-1D & 2 & [1,2] & (C$_{max}$-32):4:C$_{max}$ \\
NB-1D & 4 & [1,2] & (C$_{max}$-32):4:C$_{max}$ \\
NB-1D & 8 & [1,2] & (C$_{max}$-32):4:C$_{max}$ \\
NB-1D & 16 & [1,2] & (C$_{max}$-32):4:C$_{max}$ \\
\hline
\end{tabular}
\end{center}
\vspace{-12pt}
\end{table}

	\begin{figure*}[t]
		\centering
		\includegraphics[width=1.0\linewidth]{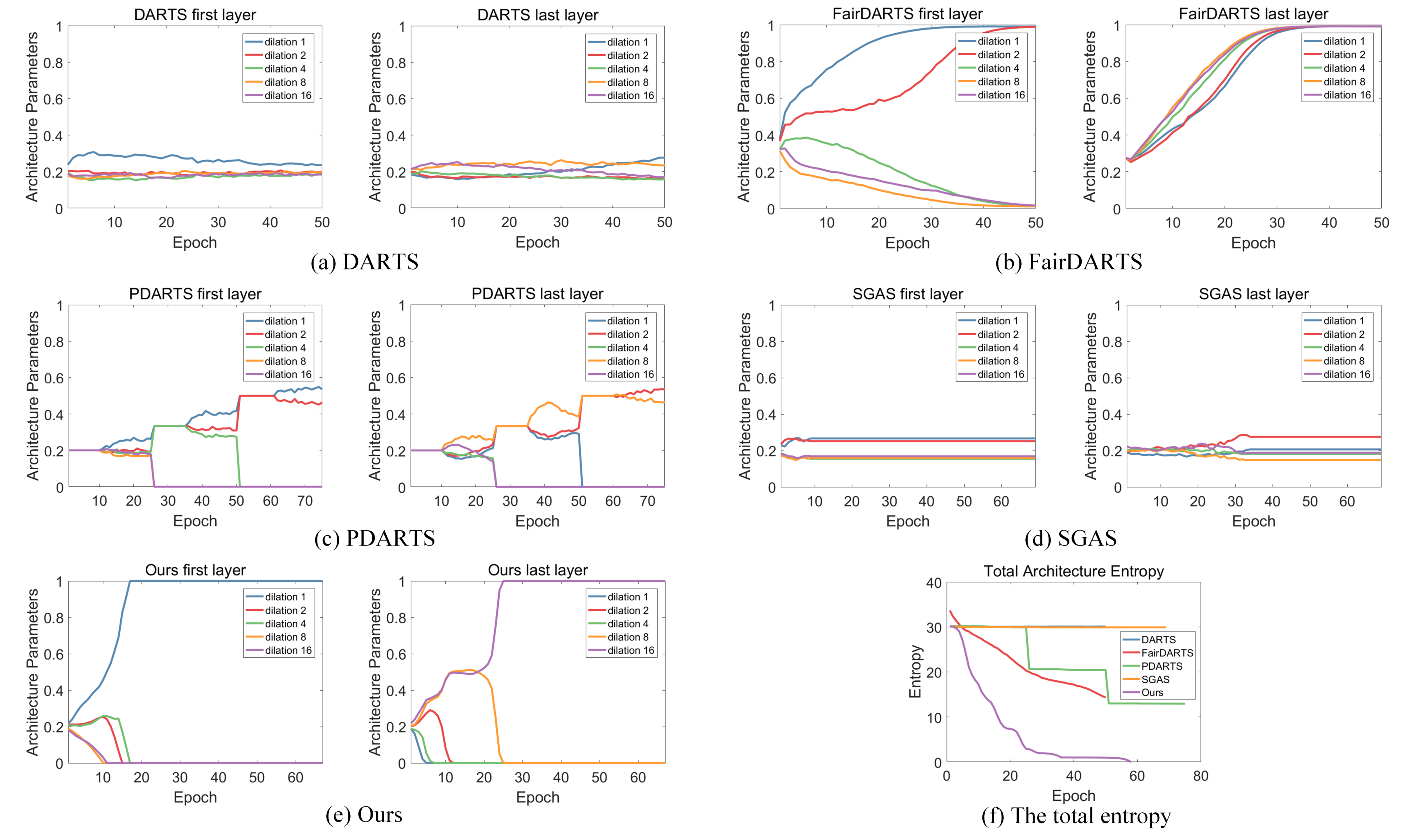}
		\caption{Illustration of the discretization gap problem. Specifically, we utilize up-to-date differentiable methods for dilation dimensional search and study the change process of architecture parameters and total architecture entropy. \bp{It can be noticed that other related methods generally keep some specified probabilities distributions for different design choices in the training process while our method can fastly lead to $[0,...,1,...,0]$ distribution. }}
		\label{fig:different_layer_prob}
		\vspace{-12pt}	
	\end{figure*}

\subsection{Implementation Details}
	
The macro and micro configurations of the searched backbone are listed in Table~\ref{tab:macro} and Table~\ref{tab:micro}. Considering the computation cost, we set the searchable depth for the 1/4 resolution stage as $\{3,4,5,6,7\}$, and for the 1/8 resolution stage as $\{6,7,8,9,10\}$, which can cover most one-branch segmentation backbones. Referring to~\cite{romera2017erfnet}, we set the searchable dilation rates for each layer as $\{1,2,4,8,16\}$. We experimentally find that the searchable spatial resolution is undesired to be too small for segmentation, thus it is set as $\{1,2\}$. In addition, the searchable channel number for $3\times 1$ and $1\times 3$ convolutions is set as $\{C_{max}-32:4:C_{max}\}$. This results in a large and complex search space, which has ${\left ( 9\times 9\right )^{\left ( 2\times 5\right ) \times \left ( 7+10\right )}}+5\times 5 \approx 2.78*10^{324}$ combinations. It is an extremely large search space which is beyond the scope of most current search algorithms. In this work, we carry out all searching experiments on Cityscapes, search on the training set and report the results on the validation set, using PyTorch 1.4.0 on one Nvidia Tesla V100 32G GPU. 

Moreover, we design a simple aggregation block for the searched backbone, to better aggregate the contextual information and spatial details, as shown in Fig.~\ref{fig:pipeline}(a). The output of each stage is fed into the aggregation block, the 1/2 resolution output is processed by an average pooling, then concatenated with the 1/4 resolution output and input to a $3 \times 3$ convolution, the 1/8 resolution output is processed by a lightweight pyramid pooling module ~\cite{zhao2017pyramid}, and finally they are concatenated to predict the segmentation map. Note that the aggregation block is only used in the retraining experiments.

\bigskip\noindent\textbf{Details for Searching.} We use half-resolution ($512 \times 1024$) images of the Cityscapes training set and its 1/8 resolution labels to search the segmentation backbone following the strategy utilized in~\cite{romera2017erfnet}. In searching comparison, we conduct all the searching process for 200 epochs. Regarding the comparison of the retraining performance, we stop the searching process once we obtain the discrete network. As for the searching hyper-parameters, we set the batch size as 6 and use Adam as the optimizer, for both architecture and network parameters. The architecture optimizer has an initial learning rate of 0.002 and \ye{a} weight decay of 0.001. The network optimizer has an initial learning rate of 0.0003 and \ye{a} weight decay of 0.0001, meanwhile the poly learning rate decay of power 0.9 is used. We experimentally find that a relatively small learning rate for network parameters is beneficial to the whole learning process. 

\bigskip\noindent\textbf{Details for Retraining.} \ye{It} should be noted that retraining means training the searched model from scratch, and it has also been widely used in the performance evaluation for searched optimal network model in NAS.  The retraining hyper-parameters on Cityscapes are the same as the searching phase, except that we only use the network optimizer and set its initial learning rate as 0.0005. For the evaluation of the searched backbone, we retrain it with half-resolution images and its 1/8 resolution labels for 200 epochs. For the evaluation of the final network, we further use an aggregation block to decode the features extracted from the searched backbone. Moreover, we use half-resolution images and labels, and employ common data augmentation strategies like color jitter, random horizontal flip, random scaling (0.375$\times$ $\sim$ 0.875$\times$) and cropping. Referring to~\cite{lin2020graph,yu2020bisenet}, the OHEM loss is also used. Following ~\cite{romera2017erfnet}, a two-stage retraining strategy is adopted to optimize the output of 1/8 resolution and original resolution for 300 epochs respectively. Note that, we do not use any extra data like coarse-annotated images or ImageNet. For the testing, the input images are first resized to half of their original resolutions and \ye{input} to our model, then we directly resize the predicted maps to the original resolution ( $1024 \times 2048$ for Cityscapes, $720 \times 960$ for CamVid, $720 \times 1280$ for BDD) to compute the mIoU.

\subsection{Results on Dilation Rate Search}
\ye{For illustrative purposes, we first present some comparison results on dilation rate dimensional search, to show the discretization gap problem and the effectiveness of the proposed method.}

\bigskip\noindent\textbf{Illustration of Discretization Gap Problem.} Specifically, we employ our method and other differentiable methods like DARTS, FairDARTS, PDARTS, SGAS for dilation dimensional search, and show the changing process of architecture parameters and the total architecture entropy. As shown in Figs.~\ref{fig:different_layer_prob}(a)-(d), all previous methods may suffer from the discretization gap problem \ye{since the architecture parameters are not distinguished sufficiently and may lead to the discrete gap in the final decision.} As shown in Fig.~\ref{fig:different_layer_prob}(e), our method has the ability to facilitate the competition of candidates with approximate weights, which can be obviously seen from the confrontation between dilation 8 (orange curve) and dilation 16 (purple one) in the last layer (right picture of Fig.~\ref{fig:different_layer_prob}(e)), and will remove candidates with low enough weights in all candidates. Thus it may properly mitigate the discretization gap problem and accelerate \ye{the} searching process. Fig.~\ref{fig:different_layer_prob}(f) further shows the change of total architecture entropy in the searching process. All other methods have a non-zero entropy that is highly related to the discretization gap problem because higher entropy means more choices with similar weights, while our method tends to smoothly reduce the total entropy to zero, indicating a rather stable searching process. 


\bigskip\noindent\textbf{Comparison to other differentiable methods.} \ye{To illustrate the efficiency and effectiveness of the proposed search scheme, we also compare the performance of the above four typical differential search methods to our proposed scheme on dilation dimensional search, and record the comparison results in Table~\ref{tab:one_dim_search}.} As shown in Table~\ref{tab:one_dim_search}, our method not only has the best search efficiency of 3.5 hours compared to about 7 hours by other methods, but also can avoid some interference caused by useless choices and be easier to find the optimal one, as evidenced by the best segmentation result of 68.06\% in terms of mIoU.

\begin{table}[!t]
\footnotesize
\centering
\caption{Comparison with other differentiable methods on dilation dimensional search.}
\label{tab:one_dim_search}
\setlength{\tabcolsep}{1.2mm}
\begin{tabular}{ccc}
\hline\noalign{\smallskip}
Method & Search Time  & Retrain mIoU(\%)  \\
\noalign{\smallskip}
\hline
\noalign{\smallskip}
DARTS~\cite{liu2018darts}  & 7.2h & 65.81 \\
FairDARTS~\cite{chu2019fair} & 7.2h & 66.57 \\
PDARTS~\cite{chen2021progressive} & 6.8h & 66.09 \\
SGAS~\cite{li2020sgas} & 6.5h & 66.59 \\
Ours & \textbf{3.5h} & \textbf{68.06}\\
\hline
\end{tabular}
\vspace{-6pt}
\end{table}

\begin{table}[t]
\footnotesize
\centering
\caption{\ye{Progressive manner study on dilation dimensional search \textbf{without} SSR loss.}}
\label{tab:progressive_manner_wo}
\begin{tabular}{ccc|c}
\hline
\multicolumn{3}{c|}{Progressive manner} &\multirow{2}{*}{Retrain mIoU(\%)}\\
\cline{1-3}
PDARTS         & SGAS     & Ours &   \\ \hline
$\checkmark$   &          &      & 66.09   \\
               & $\checkmark$ &  & \ 66.59   \\
               & & $\checkmark$  & \textbf{66.95}  \\ \hline
\end{tabular}
\vspace{-6pt}
\end{table}

\begin{table}[t]
\footnotesize
\centering
\caption{\ye{Progressive manner study on dilation dimensional search \textbf{with} SSR loss.}}
\label{tab:progressive_manner_w}
\begin{tabular}{ccc|c|c}
\hline
\multicolumn{3}{c|}{Progressive manner} &\multirow{2}{*}{SSR} &\multirow{2}{*}{Retrain mIoU(\%)}\\
\cline{1-3}
PDARTS         & SGAS     & Ours &   \\ \hline
$\checkmark$   &          &      &$\checkmark$ & 66.36   \\
               & $\checkmark$ &  &$\checkmark$ & 67.25   \\
               & & $\checkmark$  &$\checkmark$ & \textbf{68.06}  \\ \hline
\end{tabular}
\vspace{-6pt}
\end{table}

\begin{table}[t]
\centering
\footnotesize
\caption{Auxiliary loss study on dilation dimensional search with our progressive shrinking strategy.}
\label{tab:different_aux}
\begin{tabular}{ccc|c|c}
\hline
\multicolumn{3}{c|}{Auxiliary loss} &\multirow{2}{*}{Search Time} &\multirow{2}{*}{Retrain mIoU(\%)} \\
\cline{1-3}
$L_{1}$     &  $L_{2}$     & SSR & &  \\ \hline
$\checkmark$  &           &        & 5.5h & 65.47     \\
              & $\checkmark$  &    & 5.5h & 65.81  \\
             &       & $\checkmark$  & 3.5h& \textbf{68.06}  \\ \hline
\end{tabular}
\vspace{-12pt}
\end{table}

\bigskip\noindent\textbf{Influence of Different Components.} To demonstrate the proposed level-based progressive shrinking strategy, we compare it to some other progressive manners such as PDARTS and SGAS, and the corresponding results are shown in Table~\ref{tab:progressive_manner_wo} \ye{and Table~\ref{tab:progressive_manner_w}. Note that the search setting of our method without SSR loss is kept the same as SGAS, except that we remove candidates with the global lowest weights. As shown in Table~\ref{tab:progressive_manner_wo}, without SSR loss, the proposed progressive search manner can yield a performance of 66.95\% mIoU, higher than 66.09\% mIoU of PDARTS and 66.59\% mIoU of SGAS. As shown in Table~\ref{tab:progressive_manner_w}, with SSR loss, the performance of different methods has been improved consistently (i.e., 66.95\% to 68.06\% mIoU for our method, 66.09\% to 66.36\% mIoU and 66.59\% to 67.25\% mIoU for PDARTS and SGAS), while the proposed method has larger performance improvement margin.} For feasibility, PDARTS and SGAS can only be used to search for one dimensional case, while our level-based progressive shrinking strategy can be extended to joint-dimensional search. In addition, based on the progressive shrinking strategy, \ye{we further compare the proposed SSR loss with the typical $L_{1}$ and $L_{2}$ loss}~\cite{chu2019fair}, and the results are illustrated in Table~\ref{tab:different_aux}. It can be observed that when using $L_{1}$ and $L_{2}$ loss, their performances (65.47\% and 65.81\%) are inferior to our result (68.06\%), and the search time of combining $L_{1}$ and $L_{2}$ loss with our progressive manner is 5.5h, slower than ours (3.5h) with SSR loss. The possible reason is that $L_{1}$ and $L_{2}$ loss lead to multiple solutions which cannot be pruned when searching.

\subsection{Results on Joint Dimensional Search}
\begin{figure}[t]
		\vspace{-2pt}
		\centering
		\includegraphics[width=3.0in]{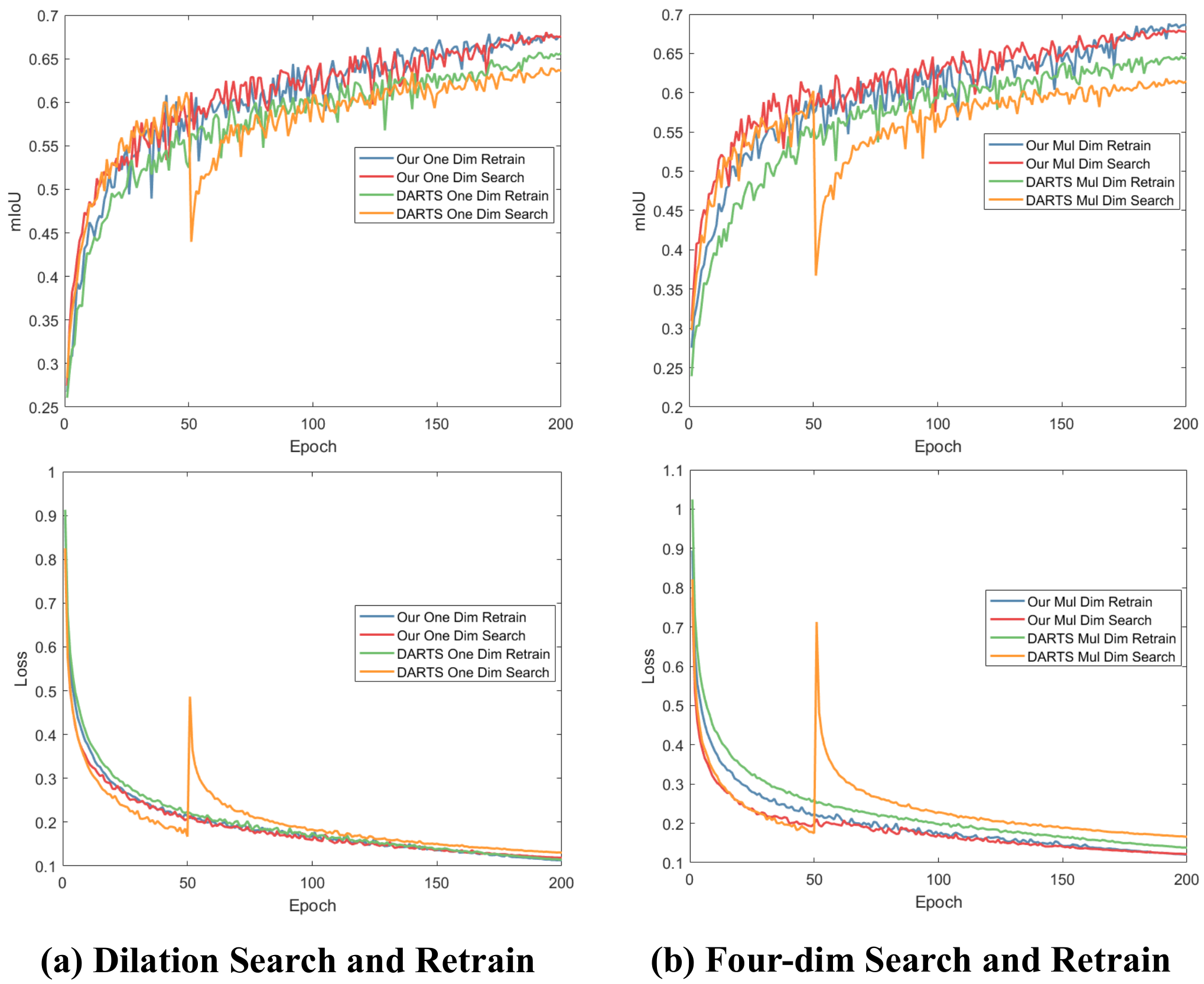}
		\caption{Searching and retraining process of our method for dilation rate and joint dimensional search. We plot the segmentation performance in terms of mIoU and the loss curve, and use the method of DARTS~\cite{liu2018darts} as a comparison.
		\ye{Searching means to complete the iterative search process and continue to train the discrete subnet. Retraining means to train the discrete network from scratch.}}
		\label{fig:gap}
		\vspace{-6pt}	
	\end{figure}

\textbf{Discretization Gap Reduction.} As mentioned before, our method can effectively reduce the gap between the supernet and its \ye{discretized} one. To depict such effects, we draw the segmentation performance in terms of mIoU and the loss curve for dilation rate ($D_{i}$) and joint ($ D_{i} + C + S + D_{e}$) dimensional search, and report them in Fig.~\ref{fig:gap}. As a comparison, we also use DARTS for dilation rate and joint dimensional search and show the searching and retraining performance. We can see that even on dilation rate dimensional search, DARTS suffers from a serious discretization gap problem. Specifically, the searching loss and mIoU (orange) curves have a dramatic performance drop when discretizing the final architecture, and there is an obvious performance gap between the searching and the retraining (green curves) process. Such \ye{discretization} gap problem is further exacerbated on joint dimensional search of DARTS. 

As we can see in Fig.~\ref{fig:gap}, on both dilation and joint dimensional search, our searching method has a smooth loss and mIoU (red) curves, which is very close to the retraining loss and mIoU (blue) curves, validating a stable and gradually approaching searching process \yep{(i.e., maximized correlation)}. \ye{While for DARTS, there is an obvious gap between the orange one and the green one for both dilation and joint dimensional search.} Thus, our method can minimize the discretization gap problem and obtain a network with better performance. 

\ye{In addition, we provide further theoretical analyses for such effects of our method. Specifically, for the bi-level optimization objective of Eq.~(\ref{eqn6}), suppose $(w^{*}, A^{*})$ is the learned optimal solution in the continuous solution space and $\hat{A}$ denotes the discrete solution. Then, based on Taylor expansion, we have:
\begin{equation} \label{eqn25}
  \begin{split}
    &L_{A}(w^{*}, \hat{A}) \\
    & \approx L_{A}(w^{*}, A^{*})+\nabla_{A} L_{A}(w^{*}, A^{*})(\hat{A}-A^{*}) \\
    &\quad +\frac{1}{2} \nabla_{A}^{2} L_{A}(w^{*}, A^{*})(\hat{A}-A^{*})^{2} \\
    &=L_{A}(w^{*}, A^{*})+\frac{1}{2} \nabla_{A}^{2} L_{A}(w^{*}, A^{*})(\hat{A}-A^{*})^{2}
  \end{split}
  \vspace{-6pt}
\end{equation}
where $\nabla_{A} L_{A}=0$ due to the optimality condition. As $L_{A}(\hat{w}, \hat{A})$ where $\hat{w}$ is the corresponding optimal weight of $\hat{A}$ is expected to be smaller than $L_{A}(w^{*}, \hat{A})$ after fine-tuning, the performance of $L_{A}(\hat{w}, \hat{A})$ will also be bounded by the second term of Eq.~(\ref{eqn25})~\cite{chen2020stabilizing}.  Mathematically, the discretization gap caused by projecting $A^{*}$ to $\hat{A}$ is directly correlated to the second term of Eq.~(\ref{eqn25}).} \yep{Furthermore, based on Eq.~\ref{eqn25}, we can get the following equation:
\begin{equation} \label{eqn26}
  \begin{split}
   (\hat{A}-A^{*})^{2} \approx \frac{1}{Z}(L_{A}(w^{*}, \hat{A})-L_{A}(w^{*}, A^{*}))
  \end{split}
  \vspace{-6pt}
\end{equation}
where $Z=\frac{1}{2}\nabla_A^2L_A(w^\ast,A^\ast)$ is a definite value as the search is done. As we can see, the value $L_A(w^\ast,\hat{A})-L_A(w^\ast,A^\ast)$ is proportional to the discretization gap, that is to say, the performance/loss of the learned supernet may not well represent the performance/loss of the derived subnet when the discretization gap is non-negligible. To solve this, the SSR loss and hierarchical progressive solution space shrinking strategy are proposed to gradually reduce the discretization gap during the search, so as to minimize the influence of projecting $A^\ast$ to $\hat{A}$ (i.e., maximize the correlation between $L_A(w^\ast,\hat{A})$ and $L_A(w^\ast,A^\ast)$). As such, we can realize the target of gradually optimizing $A^{*}$ to $\hat{A}$ so as to improve the search performance and search efficiency.}

	\begin{figure*}[t]
		\centering
		\includegraphics[width=0.95\linewidth]{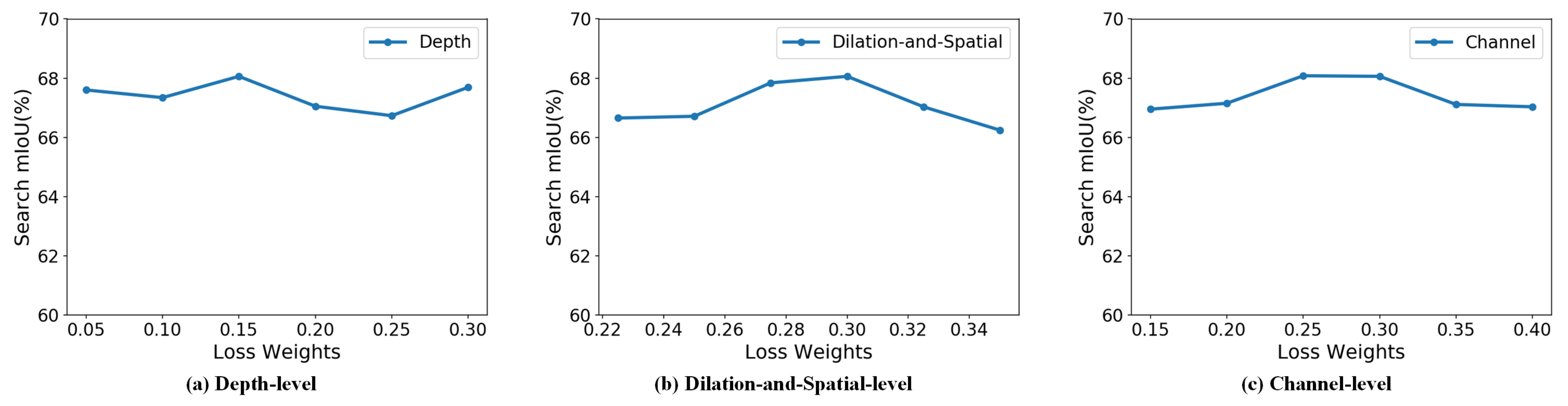}
		\caption{\ye{Ablative study on the loss weights. The optimal loss weights of Depth-level, Dilation-and-Spatial-level and Channel-level search spaces are 0.15, 0.3 and 0.3 respectively.}}
		\label{fig:loss_w}
		\vspace{-6pt}	
	\end{figure*}

\bigskip\noindent\textbf{Joint Different Dimensional Search.} To show the effects of different search spaces in the proposed scheme, we carry out comparison experiments on Cityscapes validation set. All experiments only conduct the searching process mentioned above for 200 epochs, without retraining. The results are shown in Table~\ref{tab:search_space}, \ye{where baseline1 denotes the basic network without any dimensional changes and baseline2 adds manual-designed dilation arrangement~\cite{romera2017erfnet} based on the former.} In particular, the baseline1 network has 5 normal blocks in the 1/4 resolution stage and 8 normal blocks in the 1/8 resolution stage. Each block is represented by a simplified version of the efficient residual factorized convolution~\cite{romera2017erfnet}, which contains only a set of $3 \times 1$ and $1 \times 3$ convolutions in the residual block. 

\setlength{\tabcolsep}{4pt}
\begin{table}[!t]
\begin{center}
\footnotesize
\caption{Results of joint different dimensional search, without/with a specified FLOPs constraint.}
\label{tab:search_space}
\begin{tabular}{cccc}
\hline\noalign{\smallskip}
Method & Params(M) &  FLOPs(G) & mIoU(\%)  \\
\noalign{\smallskip}
\hline
\noalign{\smallskip}
baseline1  & 0.96 & 11.2 & 59.38 \\
\ye{baseline2} & 0.96 & 11.2 & 64.84 \\
\hline
$D_{i}$ & 0.96/0.67 & 11.2/7.8 & 68.04/64.63 \\
$D_{i} + S $ & 0.96/0.96 & 9.3/7.8 & 67.79/64.91 \\
$D_{i} + C $ & 0.92/0.80 & 10.5/7.8 & 67.64/65.83\\
$D_{i} + C + S $ & 0.85/0.83 & 8.2/7.8 & 67.35/67.34 \\
$D_{i} + C + S + D_{e}$ & 0.92/0.83 & 8.5/7.8 & 68.06/67.49 \\
\hline
\end{tabular}
\end{center}
\vspace{-12pt}
\end{table}
\setlength{\tabcolsep}{1.4pt}

\setlength{\tabcolsep}{4pt}
\begin{table}[!t]
\begin{center}
\footnotesize
\caption{Search efficiency of different NAS-based real-time segmentation methods. $^\dagger$ denotes our running results.}
\label{tab:search_time}
\begin{tabular}{cccc}
\hline\noalign{\smallskip}
Method & Testbed & Search Space & Search Cost  \\
\hline
CAS~\cite{zhang2019customizable} & - & $1.0*e^{27}$ & - \\
GAS~\cite{lin2020graph} & Titan XP & $1.6*e^{63}$ & 160h \\
AutoRTNet~\cite{sun2021real} & Titan XP & $7.8*e^{34}$ & 256h\\
FasterSeg$^\dagger$~\cite{chen2019fasterseg} & V100 & $3.4*e^{55}$ & 84h\\
Ours & V100 & $2.78*e^{324}$ & 18h \\
\hline
\end{tabular}
\end{center}
\vspace{-16pt}
\end{table}
\setlength{\tabcolsep}{1.4pt}

Here, to show a comprehensive comparison, we conducted two sets of tests. One set represents the model is directly obtained with our search scheme, and the other one represents that the model is obtained by the search scheme with a specified FLOPs constraint (Since we found in our experiments that the minimum FLOPs under different search space combinations is about 8.2 G, we set 7.8G as our new constraint in the searching process to further reduce the FLOPs. Please refer to Appendix A.1 for more details). As we can see from the first set of experimental results, when applying different search space modules, the obtained models show different performances in terms of mIoU and FLOPs. Compared with depth-only search with 11.2G FLOPs, full-dimension search with 8.5G FLOPs requires much fewer FLOPs, while other results may have lower mIOU but obviously lower FLOPs. Generally speaking, the proposed search scheme can obtain better models with less FLOPs and better mIoU at the same time compared to the baselines. For our design, the full search space setting of $ D_{i} + C + S + D_{e}$ yields a small size model (0.92M) with less FLOPs (8.5G) and has the best performance of 68.06\% on Cityscapes. When applying specified FLOPs constraint in searching, consistent improvement is observed when including more designs, since the performance goes from 64.63\% when searching only $D_{i}$ to 67.49\% when simultaneously searching the four dimensions. Such a gradually increasing performance further validates the effectiveness of the proposed multi-dimensional search space design.

\begin{table*}[t]
\begin{center}
\footnotesize
\caption{Comparison results on Cityscapes test set. $^\ast$ denotes the result reported in ~\cite{lin2020graph}. TR donotes the TensorRT library. $^\dagger$ denotes our running results. Note that, we take 512$\times$1024 or 768$\times$1536 resolution images as model input, and directly resize the predicted maps to 1024$\times$2048 resolution for testing. $^\ddagger$ denotes the model is trained on 768$\times$1536 resolution.}
\label{tab:full_comp_city}
\begin{tabular}{cccccccc}
\hline\noalign{\smallskip}
Model & Input Size & Pretrained & Testbed & FLOPs(G) & Params(M) & FPS & Test mIoU(\%)  \\
\hline
ICNet~\cite{zhao2018icnet} & 1024$\times$2048 & ImageNet & Titan X & 28.3 & 26.5 &30.3  & 69.5 \\
ERFNet~\cite{romera2017erfnet}  & 512$\times$1024 & No & Titan X & 26 & 2.1 &41.7  & 68.0 \\
BiSeNet~\cite{yu2018bisenet} & 1024$\times$2048 & ImageNet & Titan XP & 121.9 & 13.4 &105.8  & 68.4 \\
DFANet A$^\ast$~\cite{li2019dfanet} & 1024$\times$1024 & ImageNet & Titan XP & 3.4 & 7.8 &52.6  & 71.3 \\
LiteSeg~\cite{Emara_2019} &512$\times$1024 & Coarse & 1080Ti & 4.9 & 4.4 & 88 &67.8 \\
BiSeNetV2$^\dagger$~\cite{yu2020bisenet} &512$\times$1024 & No & 2080Ti & 51.9 & 27.7 & 92 &72.6 \\
CAS~\cite{zhang2019customizable} & 768$\times$1536 & ImageNet & Titan XP & - & - &108.0  & 70.5 \\
GAS~\cite{lin2020graph} & 768$\times$1536 & ImageNet & Titan XP & - & - &108.4  & 71.8 \\
AutoRTNet~\cite{sun2021real} & 768$\times$1536 & ImageNet & Titan XP & - & 2.5 &110  & 72.2 \\
FasterSeg~\cite{chen2019fasterseg} & 1024$\times$2048 & No & 1080Ti+TR & 28.2 & 4.4 &163.9  & 71.5 \\
FasterSeg$^\dagger$~\cite{chen2019fasterseg} & 1024$\times$2048 & No & 2080Ti & 28.2 & 4.4 &105  & 71.5 \\
\hline
Ours & 512$\times$1024 & No & 2080Ti & 11.0 & 1.0 & 175  & \textbf{72.6} \\
Ours & 768$\times$1536 & No & 2080Ti & 24.8 & 1.0 & 85  & \textbf{74.4} \\
Ours$^\ddagger$ & 1024$\times$2048 & No & 2080Ti & 44.2 & 1.0 & 53  & \textbf{75.2} \\
\hline
\end{tabular}
\end{center}
\vspace{-12pt}
\end{table*}

\begin{table}[t]
\begin{center}
\footnotesize
\caption{Comparison results on CamVid test set. \ye{Note that, our method takes 360$\times$480 resolution images as model input, and directly resizes the predicted maps to 720$\times$960 resolution for testing.} We report the results of the original papers, while $^\dagger$ denotes our running results.}
\vspace{-8pt}
\label{tab:full_comp_cam}
\begin{tabular}{ccccccc}
\hline\noalign{\smallskip}
Model &  FLOPs(G) & Params(M)& FPS& Test mIoU(\%)  \\
\hline
ICNet~\cite{zhao2018icnet} & - & 26.5 &34.5  & 67.1 \\
BiSeNet~\cite{yu2018bisenet} & 40.3 & 13.4 &-  & 65.6 \\
DFANet A~\cite{li2019dfanet} & - & 7.8 &120  & 64.7 \\
CAS~\cite{zhang2019customizable} & - & - &169.0  & 71.2 \\
GAS~\cite{lin2020graph} & - & - &153.1  & 72.8 \\
AutoRTNet~\cite{sun2021real} & - & 2.5 & 140  & 73.5 \\
FasterSeg~\cite{chen2019fasterseg} & - & - &398.1 & 71.1 \\
FasterSeg$^\dagger$~\cite{chen2019fasterseg} & 9.2 & 4.4 &201 & 71.1 \\
\hline
Ours & 3.6 & 1.0 & 195  & \textbf{74.6} \\
\hline
\end{tabular}
\end{center}
\vspace{-16pt}
\end{table}

\begin{table}[t]
\begin{center}
\footnotesize
\caption{Comparison results on BDD val set. \ye{Note that, our method takes 360$\times$640 resolution images as model input, and directly resizes the predicted maps to 720$\times$1280 resolution for testing.} We report the results of the original papers, while $^\dagger$ denotes our running results .}
\label{tab:full_comp_bdd}
\begin{tabular}{ccccccc}
\hline\noalign{\smallskip}
Model &  FLOPs(G) & Params(M)& FPS& Val mIoU(\%)  \\
\hline
DRN-D-22~\cite{2017Dilated} & - & - &21  & 53.2 \\
DRN-D-38~\cite{2017Dilated} & - & - &13  & 55.2 \\
FasterSeg~\cite{chen2019fasterseg} & - & - &318 & 55.1 \\
FasterSeg$^\dagger$~\cite{chen2019fasterseg} & 12.4 & 4.4 &196 & 55.1 \\
\hline
Ours & 4.9 & 1.0 & 188  & \textbf{59.1} \\
\hline
\end{tabular}
\end{center}
\vspace{-16pt}
\end{table}

\bigskip\noindent\textbf{Search Efficiency.} To show the search efficiency of the proposed search scheme, we further compare our joint four-dimensional search with \ye{existing} NAS-based real-time segmentation methods, as shown in Table~\ref{tab:search_time}. We can see that our method is conducted on the largest search space (about $2.78*e^{324}$) but achieve the fastest search (about 18h), denoting the high search efficiency of the proposed joint-dimensional search framework.

\ye{\bigskip\noindent\textbf{Ablation study on the loss weights.} The detailed ablative experiments about the loss weights for different levels are shown in Fig.~\ref{fig:loss_w}. We experimentally find that the optimal loss weights of Depth-level, Dilation-and-Spatial-level and Channel-level search spaces are 0.15, 0.3 and 0.3 respectively. Since there exist three levels, we perform the ablation study for one level with others fixed. For the ablation of Depth level, different weights show slightly different results, and the best result is obtained when the weight is 0.15. For the ablation of Dilation-and-Spatial level and Channel level, we can always get a good result in a range of weights, about 0.28 to 0.30 for Dilation-and-Spatial level and 0.25 to 0.30 for Channel level. There are two possible factors that affect the weights of different levels. On one hand, different levels of search space have different influences on the search process. On the other hand, different levels of search space have different sizes.}

\subsection{Real-time Semantic Segmentation}
In this section, we show the results of real-time semantic segmentation on Cityscapes, CamVid and BDD to verify the effectiveness of the best model searched by the proposed joint-dimensional search scheme.

\bigskip\noindent\textbf{Results on Cityscapes.} 
We compare the searched model to the state-of-the-art (SOTA) models such as ICNet~\cite{zhao2018icnet}, ERFNet~\cite{romera2017erfnet}, BiSeNet~\cite{yu2018bisenet}, DFANet~\cite{li2019dfanet}, LiteSeg~\cite{Emara_2019}, BiSeNetV2~\cite{yu2020bisenet}, CAS~\cite{zhang2019customizable}, GAS~\cite{lin2020graph}, AutoRTNet~\cite{sun2021real} and FasterSeg ~\cite{chen2019fasterseg} in this field. The comparison results are illustrated in Table~\ref{tab:full_comp_city}. Most results for the SOTA algorithms in Table~\ref{tab:full_comp_city} are from their reported results in the original papers. From this table, it can be observed that the model searched with the full settings at the same time can yield a very high running speed of 175 FPS with a promising mIoU of 72.6\% and a small size of the whole model (1M), which is impressive compared to the SOTA methods. We obtain these results by training the searched model on 512$\times$1024 resolution. For results on 768$\times$1536 resolution, we train the searched model as that of 512x1024 resolution, the results are 74.4\% test mIoU, 85 FPS, 1.0M parameters and 24.8G FLOPs. For results on 1024x2048 resolution, we directly use the trained model on 768x1536 resolution for testing, the results are 75.2\% test mIoU, 53 FPS, 1.0M parameters and 44.2G FLOPs. These promising results further demonstrate the superior of the searched model. Note that, the predicted 512$\times$1024 or 768$\times$1536 resolution maps are directly resized to 1024$\times$2048 resolution for testing, and the accuracy is achieved without using any coarse-annotated images or ImageNet data.

\bigskip\noindent\textbf{Results on BDD.} In addition, we also transfer the searched model on Cityscapes to perform comparison tests on the BDD dataset, and show the results in Table~\ref{tab:full_comp_bdd}. Note that only a few works have considered real-time segmentation on BDD. As we can see, the model searched by the proposed scheme still maintains an impressive performance of 59.1\% mIoU, 188 FPS, 1.0M parameters and 4.9G FLOPs, which further verify the promising performance of the searched model and the encouraging transferability of the proposed joint-dimensional search scheme.

\bigskip\noindent\textbf{Results on CamVid.} We further conduct comparison tests on the CamVid dataset by applying the searched model in the above section and show the results in Table~\ref{tab:full_comp_cam}. In this table, we mainly compare FLOPs, model size, FPS and performance in terms of mIoU. It may be noticed that the searched model obtained by our joint-dimensional search method also achieves the new SOTA on CamVid dataset, with a promising performance of 74.6\% mIoU, 195 FPS, 1.0M parameters and 3.6G FLOPs. These superior results verify the transferability of the proposed NAS framework.

\begin{table}[t]
\footnotesize
\centering
\caption{Ablation study for retraining the searched model on Cityscapes validation set.}
\label{tab:AS_retraining}
\begin{tabular}{cccc|c}
\hline\noalign{\smallskip}
 Search & Retrain & Aggregation block & Training skills & mIoU(\%)  \\
\hline
 $\checkmark$ &  &  &  & 68.06 \\
 & $\checkmark$ &  &  & 68.70 \\
 & $\checkmark$ & $\checkmark$ &  & 69.55 \\
 & $\checkmark$ & $\checkmark$ & $\checkmark$ & 73.67 \\
\hline
\end{tabular}
\vspace{-8pt}
\end{table}

\begin{table}[!t]
\footnotesize
\centering
\caption{\yep{Comparison between SSR and IE loss on dilation dimensional search. We set the total epoch number to 200. Search Time means the running time until the total architecture entropy is 0 or the epoch is 200. Search mIoU is shown only when the learned architecture distribution has been optimized to the discrete solution. The original loss weight of Dilation-level search space is 0.3 and denoted as $\rm W$ here.}}
\label{tab:IE_d}
\setlength{\tabcolsep}{1.05mm}
\begin{tabular}{cccccc}
\hline
\begin{tabular}[c]{@{}c@{}}Loss\\ (weight)\end{tabular} & \begin{tabular}[c]{@{}c@{}}Total \\ Time\end{tabular} & \begin{tabular}[c]{@{}c@{}}Search \\ Time\end{tabular} & \begin{tabular}[c]{@{}c@{}}FLOPs\\ (G)\end{tabular} & \begin{tabular}[c]{@{}c@{}}Search\\ mIoU(\%)\end{tabular} & \begin{tabular}[c]{@{}c@{}}Retrain\\ mIoU(\%)\end{tabular} \\ \hline
IE($\rm W$)                                               & 14.2h                                                 & 14.2h                                                  & 11.2                                                & -                                                         & 67.54                                                      \\
IE($\rm 2W$)                                              & 9.0h                                                    & 5.0h                                                   & 11.2                                                & 67.81                                                     & 67.97                                                      \\
IE($\rm 3W$)                                                   & 7.8h                                                   & 3.3h                                                    & 11.2                                                 & 67.64                                                         & 67.72                                                      \\
SSR($\rm W$)                                              & 7.0h                                                    & 3.5h                                                   & 11.2                                                & 68.04                                                     & 68.06                                                      \\ \hline
\end{tabular}
\vspace{-4pt}
\end{table}

\begin{table}[!t]
\scriptsize
\centering
\caption{\yep{Comparison between SSR and IE loss on four dimensional search. We set the total epoch number to 200. Search Time means the running time until the total architecture entropy is 0 or the epoch is 200. Search mIoU is shown only when the learned architecture distribution has been optimized to the discrete solution.The original loss weights of Depth-level, Dilation-and-Spatial-level and Channel-level search spaces are 0.15, 0.3 and 0.3 respectively, and denoted as $\rm W_1$, $\rm W_2$ and $\rm W_3$ here.}}
\label{tab:IE_f}
\setlength{\tabcolsep}{1.05mm}
\begin{tabular}{cccccc}
\hline
\begin{tabular}[c]{@{}c@{}}Loss\\ (weight)\end{tabular} & \begin{tabular}[c]{@{}c@{}}Total \\ Time\end{tabular} & \begin{tabular}[c]{@{}c@{}}Search \\ Time\end{tabular} & \begin{tabular}[c]{@{}c@{}}FLOPs\\ (G)\end{tabular} & \begin{tabular}[c]{@{}c@{}}Search\\ mIoU(\%)\end{tabular} & \begin{tabular}[c]{@{}c@{}}Retrain\\ mIoU(\%)\end{tabular} \\ \hline
IE($\rm W_1$, $\rm W_2$, $\rm W_3$)                                                      & 34.0h                                                   & 34.0h                                                    & 7.3                                                 & -                                                         & 66.46                                                      \\
IE($\rm 2W_1$, $\rm 2W_2$, $\rm 2W_3$)                                              & 23.9h                                                    & 23.9h                                                   & 8.9                                                & -                                                     & 67.33                                                      \\
IE($\rm 3W_1$, $\rm 3W_2$, $\rm 3W_3$)                                                   & 19.4h                                                   & 13.3h                                                    & 10.2                                                 & 66.83                                                         & 67.62                                                      \\
IE($\rm 2W_1$, $\rm 3W_2$, $\rm 4W_3$)                                                   & 18.6h                                                   & 11.9h                                                    & 8.3                                                 & 67.66                                                         & 68.10                                                      \\
SSR($\rm W_1$, $\rm W_2$, $\rm W_3$)                                                  & 18h                                                   & 11h                                                    & 8.5                                                 & 68.06                                                     & 68.70                                                      \\ \hline
\end{tabular}
\vspace{-4pt}
\end{table}

	\begin{figure}[t]
		\centering
		\includegraphics[width=3.4in]{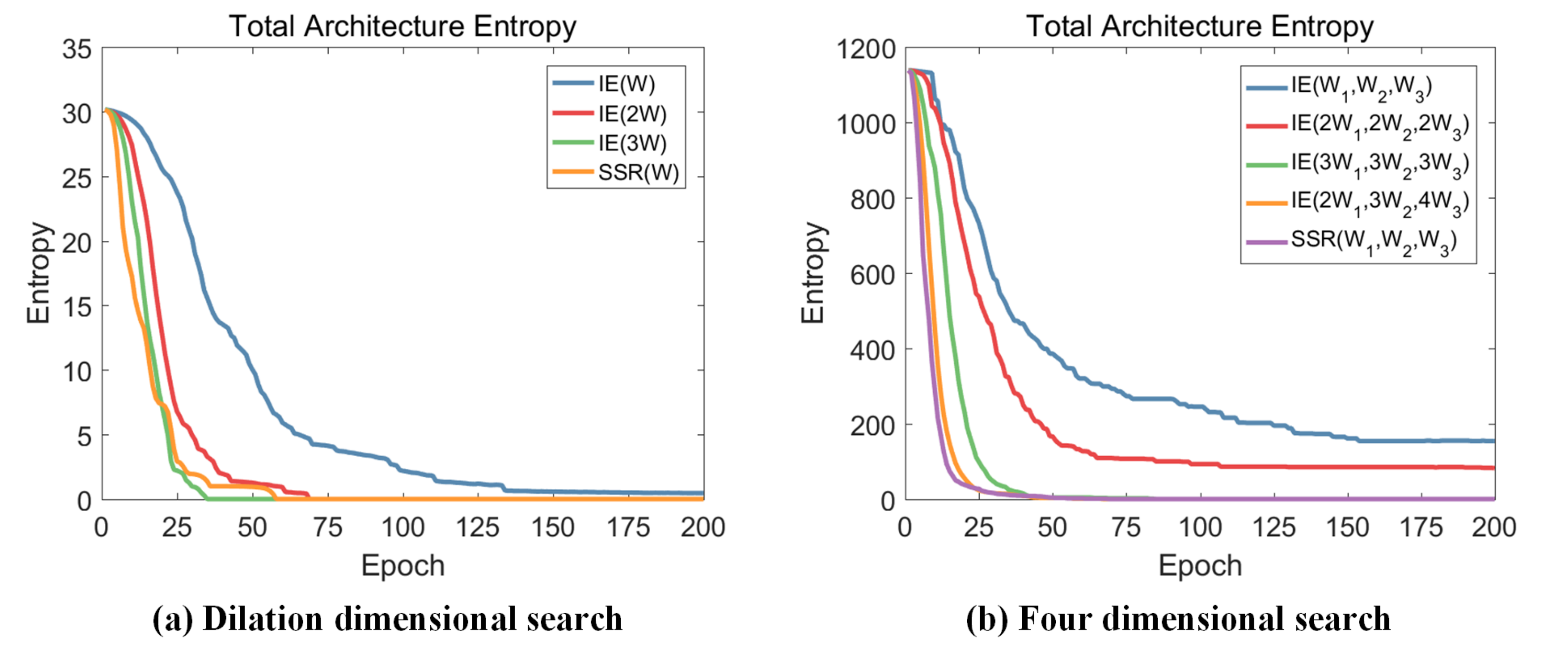}
		\vspace{-18pt}
		\caption{\yep{Comparison between SSR and IE loss on dilation and four dimensional search.}}
		\label{fig:IE}
		\vspace{-6pt}	
	\end{figure}

\subsection{Ablation Study for Retraining.} For conveniently studying joint-dimensional search and the discretization gap problem, we conduct searching experiments without using any training skills and limit the maximum epochs. To show the effects of different components in the retraining process that are taken by most works in the field of NAS, we further carry out ablation \ye{studies} for retraining on Cityscapes and show the results in Table~\ref{tab:AS_retraining}. The baseline is the model's performance after \ye{the} direct search of 200 epochs (val 68.06\%). When retraining the searched model for 200 epochs, the performance gets slightly better (val 68.70\%). With the aggregation block, we can obtain higher accuracy (val 69.55\%) and higher resolution of segmentation results. When using common training skills (data augmentation, OHEM loss, iterative training for 300 epochs), the performance is greatly boosted (val 73.67\%). As for the results of 512$\times$1024 resolution on test set (test 72.6\%), we use the training set and validation set for training following the practices in SOTA works. 

\subsection{Feature and Results Visualization}

\begin{figure*}[t]
	\centering
	\includegraphics[width=0.95\linewidth]{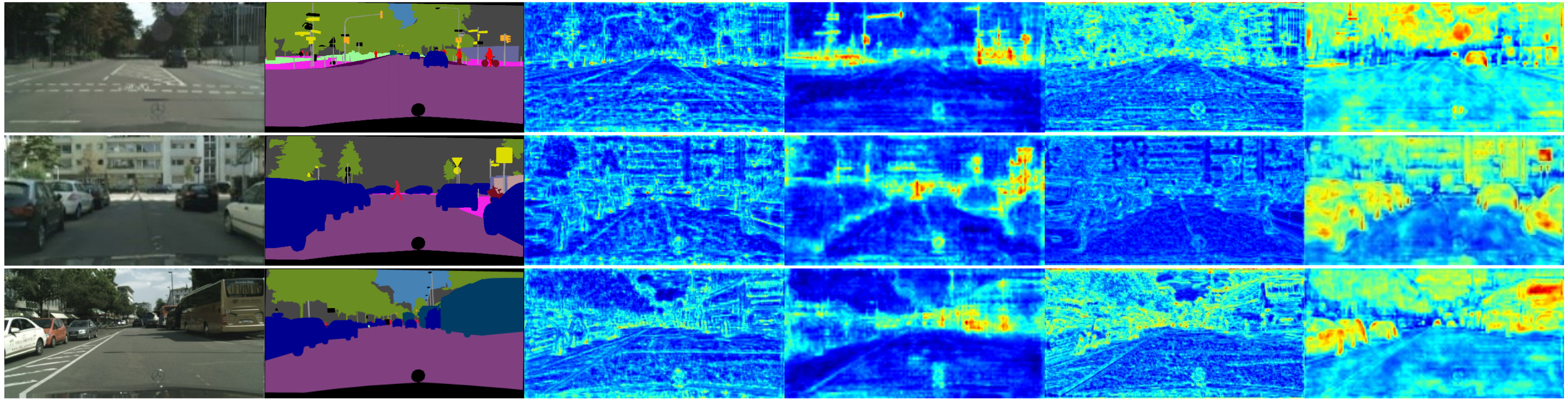}
	\caption{ Feature map visualization. From left to right are images, ground truths, output feature map of 1/4 searched block, 1/8 searched block, 1/4 baseline block, 1/8 baseline block.(Best viewed in color)}
	\label{fig:heat_map}
	\vspace{-6pt}	
\end{figure*}

\textbf{Feature Map Visualization.} 
\ye{To illustrate the better feature description ability of the model obtained by the proposed joint-dimensional search method, we visualize the output feature maps of 1/4 and 1/8 searched blocks.} As shown in Fig.~\ref{fig:heat_map}, better spatial details (\ye{e.g.,} edge) and contextual information (\ye{e.g.,} person) are captured by the searched blocks, which are achieved by the joint search of dilation rates, feature spatial resolutions, depths and channels. 

\begin{figure*}[t]
	\centering
	\includegraphics[width=0.95\linewidth]{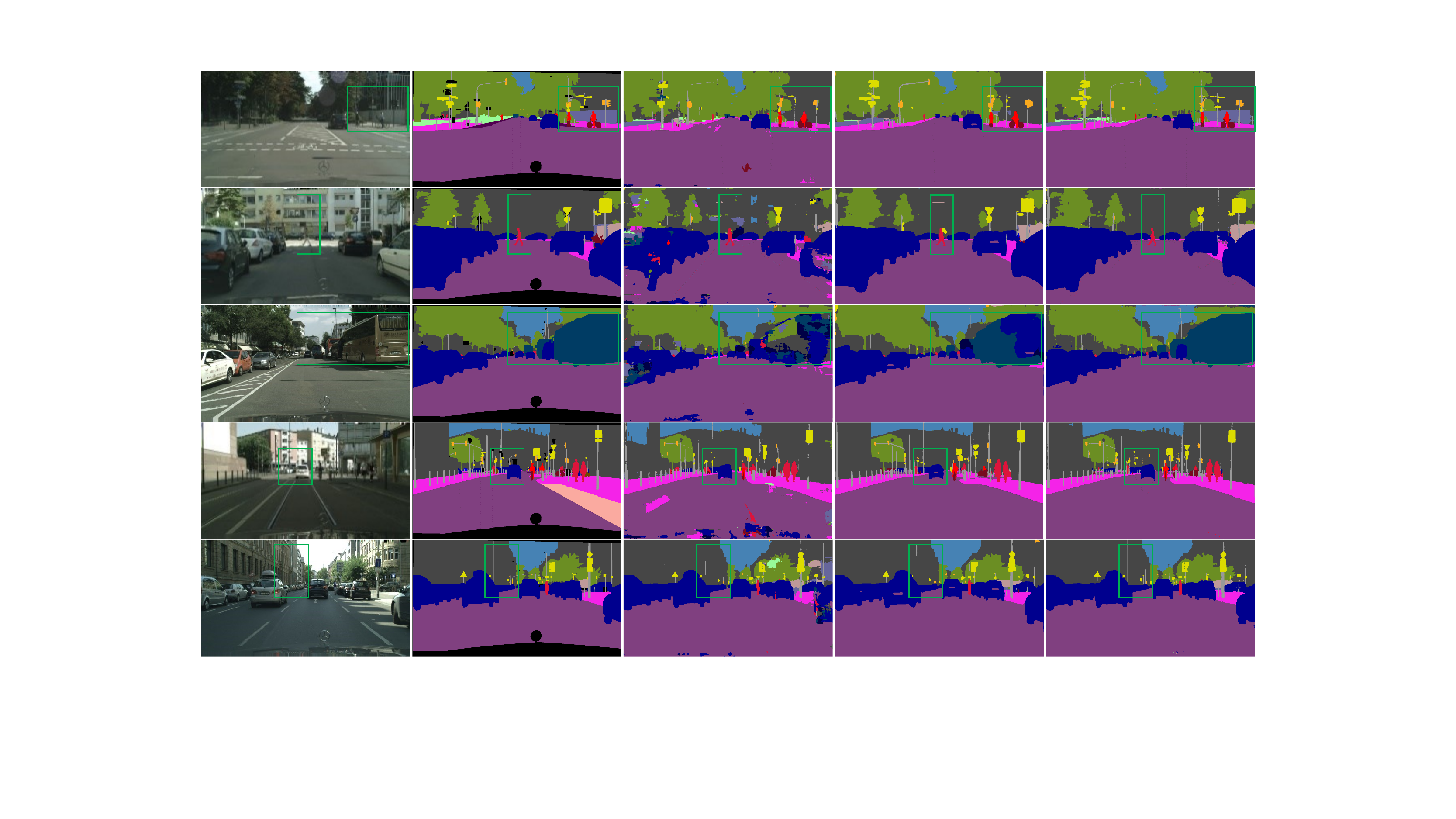}
	\caption{Segmentation visualization on Cityscapes validation set. Columns from left to right depict example images, ground truths, and results of baseline1, FasterSeg~\cite{chen2019fasterseg}, our method. Some error-prone regions are highlighted with green rectangles (Best viewed in color).}
	\label{fig:val set}
	\vspace{-12pt}	
\end{figure*}

\bigskip\noindent\textbf{Segmentation Visualization.} We conduct segmentation visualization on the Cityscapes dataset. Some visualized segmentation results of the validation set are shown in Fig.~\ref{fig:val set}. \ye{By comparing the results in the fifth column to those of the third column, we can find that} our method of joint-dimensional search can significantly improve the segmentation results of baseline1 network. As shown in the fourth column, FasterSeg may ignore some important categories when multiple categories exist in the same scene (highlighted part of the \ye{first image}), or produce some false segmentation (highlighted part of the second and third images), or miss some details and small targets (highlighted part of the fourth and fifth images), while our method can better handle these problems. 

\begin{figure*}[t]
	\centering
	\includegraphics[width=0.95\linewidth]{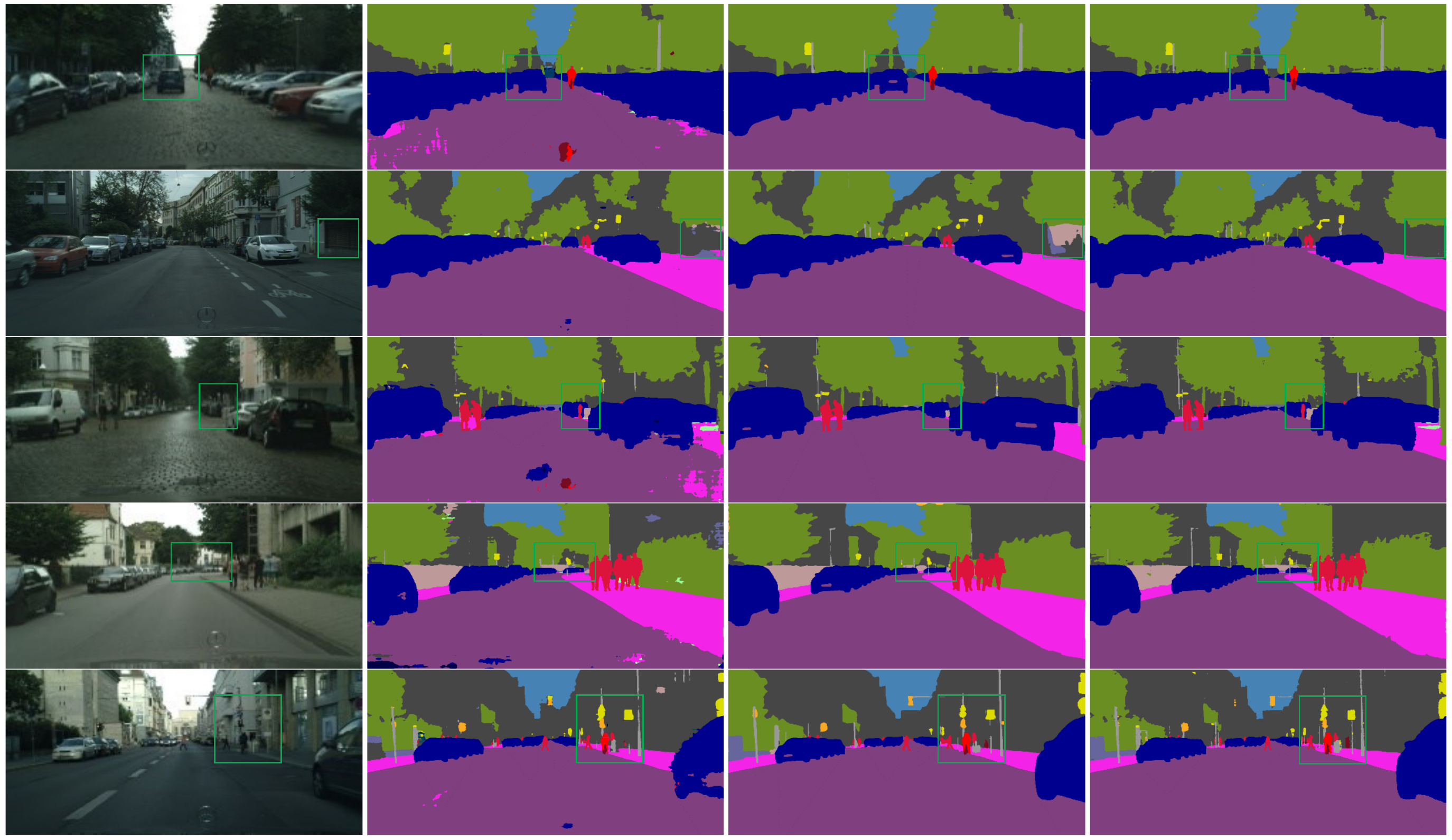}
	\caption{Segmentation visualization on Cityscapes test set. Columns from left to right depict example images, and results of baseline1, FasterSeg~\cite{chen2019fasterseg}, our method. Some error-prone regions are highlighted with green rectangles (Best viewed in color).}
	\label{fig:test set}
	\vspace{-12pt}
\end{figure*}

We also show some visualized segmentation results of the test set in Fig.~\ref{fig:test set}. Our method is still superior to baseline1 and FasterSeg networks. Specifically, FasterSeg may falsely predict the house at the end of the road (highlighted part of the first image) and the door of the building at the corner (highlighted part of the second image), or neglect the person by the car (highlighted part of the third image), or disregard the overall structural information of some categories (highlighted part of the fourth and fifth images). Generally speaking, two indispensable factors, \ye{that is contextual information and spatial details}, are needed to overcome these difficult examples. In this paper, we utilize the proposed joint-dimensional search framework to realize more contextual information and spatial details preservation, thus we can achieve a more promising segmentation performance.

\section{Discussion}
\yep{In addition, we further employ the information entropy (IE, namely $p_i log(p_i)$) as the regularization loss and increase the weight of IE loss, on both dilation and multiple dimensional search. The comprehensive comparisons with the proposed SSR loss are shown in Fig.~\ref{fig:IE}, Table~\ref{tab:IE_d} and Table~\ref{tab:IE_f}.

We demonstrate that SSR loss has a stronger sparsification effect than IE loss. Experimentally, from Fig.~\ref{fig:IE}, we can see that, when IE loss and SSR loss have the same weight, the search process with IE loss does not well converge to a fully discrete solution (i.e., total architecture entropy is 0), especially for four-dimensional search. When the weight value of IE loss is twice that of SSR loss, the total architecture entropy on dilation dimensional search with IE loss is gradually reduced to 0, while the total architecture entropy on four-dimensional search with IE loss still does not converge to a fully discrete solution. When the weight value of IE loss is three times that of SSR loss, the total architecture entropy on both dilation- and four- dimensional search with IE loss are gradually reduced to 0. In summary, IE loss does have similar effects as the proposed SSR loss, that is minimizing the total architecture entropy so as to relieve the discretization gap. Differently, SSR loss has a stronger sparsification effect than IE loss since the weight value of IE Loss needs to be several times that of SSR loss to achieve similar results, especially for high and multiple dimensional search. Theoretically, we demonstrate that SSR loss can take the same effect as L0-norm regularization (as evidenced by Section 4), thus have the best sparsification effect.

From Table~\ref{tab:IE_d} and Table~\ref{tab:IE_f}, we can see that, on both dilation- and four- dimensional search, we can use the search performance to approximate the retrain performance as long as the discretization gap is well handled, and relieving the discretization gap better can improve the performance of searched architecture better. When IE loss and SSR loss have the same weight, the time cost of the search process with IE loss is much higher than that with SSR loss, and the performance of the searched architecture via IE loss is inferior to that via SSR loss, on both dilation- and four- dimensional search. On dilation dimensional search, when the weight value of IE loss is two or three times that of SSR loss, we could obtain a similar entropy curve as SSR loss thus a similar search time and performance. Similar results can be found on four dimensional search when the weight values of IE loss on depth, dilation-and-spatial and channel-level search spaces are two, three and four times that of SSR loss respectively. Note that, all the results in Table~\ref{tab:IE_d} and Table~\ref{tab:IE_f} are much higher than DARTS search results on dilation- (65.81 retrain mIoU) and four- (64.61 retrain mIoU) dimensional. In summary, all these experimental results validate the effectiveness of the proposed idea of solution space regularization, including solution space regularization loss and solution space shrinking strategy. 

It is worth mentioning that, using information entropy (IE) as the regularization loss has not been explored in previous works, and the perspective of Solution Space and the idea of Solution Space Regularization are introduced for the first time in this paper. In other words, IE loss is also an implementation investigated in this paper for achieving the goal of encouraging the learned architecture distribution to gradually converge to the final discrete solution. Along this direction, designing new solution space regularization losses and comparing them are two promising directions worth studying in the future.
}

\section{Conclusions}
In this work, for semantic segmentation, we have advanced a differentiable joint architecture search method that finds the optimal network depths, channels, dilation rates and feature spatial resolutions, realizing more contextual information and spatial details preservation. To overcome the issue of the extremely large search space and \ye{the} intractable discretization gap problem, a novel Solution Space Regularization (SSR) loss is proposed to make the learned architecture parameters closer to the final discrete ones for discretization gap minimization, and a new hierarchical and progressive solution space shrinking strategy is introduced to improve the search efficiency and reduce the resource consumption. We theoretically demonstrate that the optimization of SSR loss is equivalent to $L_{0}$-norm regularization, and we experimentally show that the discretization gap problem can be well handled. Comprehensive tests on typical benchmark datasets demonstrate the superior performance of the whole method with high speed, light size and better accuracy at the same time.

\bibliographystyle{spmpsci}      
\bibliography{nas_seg}

\end{document}


\title{Supplementary File for `Efficient Joint-Dimensional Search with Solution Space Regularization for Real-Time Semantic Segmentation' 
}


\author{Peng Ye$^\dagger$   \and
        Baopu Li$^\dagger$  \and
        Tao Chen \textsuperscript{*} \and
        Jiayuan Fan  \and
        Zhen Mei  \and
        Chen Lin  \and
        Chongyan Zuo  \and
        Qinghua Chi  \and
        Wanli Ouyang
}

\date{Received: date / Accepted: date}

\maketitle

In this supplementary material, we intend to show more experimental details and results, including FLOPs constraint and searched structure, and give more discussions, including insights from experiments and possible future works.

\section{More Details and Results}
\subsection{FLOPs Constraint}
As mentioned in Section 5.4 of the experiments in the main text, we apply FLOPs constraint to guide the search process for a more intuitive comparison of different search space modules. Specially, we use 
certain FLOPs (7.8G in this work) as budget regularization and optimize it via gradient descent.

Referring to~\cite{guo2020dmcp}, we firstly compute the expected output channel number of each convolution as:
\begin{equation} \label{eqn2}
  \begin{split}
    &E_{conv}^{i,j,r,s} = sum(\sum_{\gamma_k^{i,j,r,s}} 
    \sigma(\gamma_k^{i,j,r,s}) M_k^{i,j,r,s}),   \\
    &\gamma_k^{i,j,r,s}\in {Arch(C)}
  \end{split}
\end{equation}
where $\gamma_k^{i,j,r,s}$ means the {channel-level} architecture parameters, $M_k$ denotes the zero-one masks. $E_{conv}^{i,j,r,s}$ is the expected output channel number of $conv$, which has dilation rate $r$ and spatial resolution $s$ and belongs to layer $j$ of stage $i$.

After obtaining expected output channel number of $3 \times 1$ and $1 \times 3$ convolution, that is, $E_{3 \times 1}^{i,j,r,s}$ and $E_{1 \times 3}^{i,j,r,s}$, the expected FLOPs of the operator with  dilation rate $r$ and spatial resolution $s$  can be thus computed as:
\begin{align}
&E_{r,s}^{i,j}=\begin{cases}
E_{kernel}^{i,j,r,s} \times H_I \times W_I&\text{$s=1$}\\
(E_{kernel}^{i,j,r,s}+s^2 \times E_{in}^{i,j,r,s})  \times \frac{H_I}{s} \times \frac{W_I}{s}&\text{$other$}
\end{cases}  \nonumber \\[12pt]
&E_{kernel}^{i,j,r,s}=(3 \times E_{in}^{i,j,r,s} + 1) \times E_{1 \times 3}^{i,j,r,s} \\
&\quad\quad\quad\quad+(3 \times E_{1 \times 3}^{i,j,r,s} + 1) \times E_{3 \times 1}^{i,j,r,s} \nonumber
\end{align}
where $E_{in}^{i,j,r,s}$ means expected input channels. $H_I$ and $W_I$ denote input feature height and width respectively.

Given that each layer has multiple operators with different dilation rates and spatial changes, the dilation-and-spatial-level expected FLOPs can be computed as:
\begin{equation} \label{eqn_di_s_search}
  \begin{split}
  &E_{j}^{i} = \sum\limits_{\beta^{i,j}_{m,n}} 
  \sigma(\beta^{i,j}_{r,s}) E_{r,s}^{i,j}, \\
  &\beta^{i,j}_{r,s}\in Arch(D_{i},S)
  \end{split}
\end{equation}

Considering that each stage has several basic layers, we denote its  cumulative FLOPs as $E_{basic}$, thus the depth-level expected FLOPs can be computed as:
\begin{equation}
E_i = E_{basic}+\sum\limits_{\alpha_j^i} \sigma(\alpha_j^i)
E_j^i,\quad \alpha_j^i\in Arch(D_{e})
\end{equation}

Because we search on two stages of segmentation backbone, the expected FLOPs of the learned model is computed as:
\begin{equation}
E_{F} = \sum{E_i}, \quad for \ i=1, 2
\end{equation}

Given target FLOPs $T_{F}$, the differentiable FLOPs constraint can be formulated as:
\begin{align}
&loss_{F}=\begin{cases}
\lambda log(\left |E_{F} - T_{F} \right |) &\text{$E_{F} < \delta T_{F}$}\\
0 &\text{$other$}\\
\lambda log(\left |E_{F} - \delta T_{F} \right |) &\text{$E_{F} > T_{F}$}
\end{cases} 
\end{align}
where $\delta$ is single side tolerance ratio which is used to make sure expected FLOPs lower around target FLOPs, and we set it as 0.95 in this paper. $\lambda$ is the relative weight and is set as 0.01. We combine this FLOPs loss with the architecture parameter loss to search a network with desired FLOPs.

\subsection{Searched Structure} 



As shown in Section 5.4 of the experiments in the main text, Our full search space setting of $ D_{i} + C + S + D_{e}$ yields a small size model (0.92M) with less FLOPs (8.5G) and has the best performance of 68.06\% mIoU on Cityscapes. In Table~\ref{tab:searched_blocks}, we show the network depths, dilation rates, feature spatial resolutions and channels obtained by our joint-dimensional search scheme. We have the searched number of output channels for $3 \times 1$ and $1 \times 3$ convolution respectively. From this Table, we can see that all the four dimensions jointly contribute to the final architecture, suggesting the importance of joint-dimensional search. Moreover, from the channel dimension, it may be observed that $3 \times 1$ and $1 \times 3$ convolution of each block tend to present different channel numbers. From the dilation dimension, we can see that bigger dilation rates incline to occur in the deeper blocks of the searched network, which is consistent with the rules found in manual-designed networks~\cite{chen2017rethinking}.

\setlength{\tabcolsep}{4pt}
\begin{table}[!t]
\begin{center}
\footnotesize
\caption{Blocks obtained based on our joint-dimensional search scheme. We conduct the network depth, dilation rate, spatial resolution and channel search for the 1/4 and 1/8 resolution stages on the segmentation backbone.}
\label{tab:searched_blocks}
\begin{tabular}{ccccc}
\hline\noalign{\smallskip}
Stage & Depth$(D_{e})$ & Dilation$(D_{i})$ & Spatial$(S)$ & Channels$(C)$\\
\noalign{\smallskip}
\hline
\noalign{\smallskip}
1/4 & 1 & 1 & 1 & (64, 64)\\
1/4 & 2 & 1 & 1 & (60, 52)\\
1/4 & 3 & 1 & 1 & (64, 44)\\
1/4 & 4 & 1 & 1 & (60, 48)\\
1/8 & 1 & 4 & 2 & (112, 124)\\
1/8 & 2 & 2 & 1 & (128, 120)\\
1/8 & 3 & 1 & 1 & (108, 116)\\
1/8 & 4 & 4 & 1 & (124, 124)\\
1/8 & 5 & 8 & 1 & (120, 120)\\
1/8 & 6 & 2 & 1 & (124, 128)\\
1/8 & 7 & 8 & 1 & (120, 124)\\
1/8 & 8 & 16 & 1 & (100, 124)\\
1/8 & 9 & 2 & 2 & (96, 124)\\
\hline
\end{tabular}
\end{center}
\vspace{-6pt}
\end{table}
\setlength{\tabcolsep}{1.4pt}

\section{More Discussions}
\subsection{Insights from Experiments}

As can be seen from Table 8, Table 9 and Table 10 in the main text, NAS-based methods have more obvious advantages in terms of mIoU and FPS than manual designed methods, suggesting the strong ability of NAS in optimizing both accuracy and resource utilization in the same time. Moreover, low FLOPs and parameters seem to be not positively correlated with high FPS, especially in segmentation methods using depthwise separable convolution~\cite{li2019dfanet}. Though it is difficult, our searched model can yield a very high running speed with a promising mIoU and a small size of the whole model at the same time. The main reason is that the model jointly searched on different dimensions can extract more contextual information and spatial details.

\subsection{Possible Future Works}

In general, there are multiple objects of different shapes in an image, for example, cars are horizontal but people are vertical. Thus, making the network have task- and scene-specific 
different receptive fields along different directions with NAS may be a valuable research direction. For example, we may apply NAS to search different operators 
with different receptive fields
such as 3$\times$1 and 1$\times$3 with different dilation rates 
or 1D pooling with different size along different directions.
In addition, joint-dimensional search may also be applied to other tasks such as object detection and tracking, image generation and enhancement, and so on, since  joint-dimensional search may also provide  better features or representation capacity for the above tasks.

\bibliographystyle{spmpsci}      
\bibliography{nas_seg}

\clearpage




%
%


